\documentclass[acmsmall,screen]{acmart}
\AtBeginDocument{%
  \providecommand\BibTeX{{%
    \normalfont B\kern-0.5em{\scshape i\kern-0.25em b}\kern-0.8em\TeX}}}
\setcopyright{acmcopyright}
\copyrightyear{2022}
\acmYear{2022}

\acmJournal{CSUR}

\renewcommand\footnotetextcopyrightpermission[1]{} 
\setcopyright{none}

\usepackage{subfigure}
\usepackage{supertabular}
\usepackage{tabu}
\usepackage{booktabs,tabularx} %
\usepackage{multirow}

\usepackage{makecell}
\usepackage{array}

\begin{document}

\title{A Survey of Generative AI for Intelligent Transportation Systems: Road Transportation Perspective}

\author{Huan~Yan}
\email{yanhuanthu@gmail.com}
\author{Yong~Li}
\email{liyong07@tsinghua.edu.cn}
\affiliation{%
  \institution{Beijing National Research Center for Information Science and Technology (BNRist), Department of Electronic Engineering, Tsinghua University}
  \country{China}
  }

\begin{abstract}

 Intelligent transportation systems are vital for modern traffic management and optimization, greatly improving traffic efficiency and safety. With the rapid development of generative artificial intelligence (Generative AI) technologies in areas like image generation and natural language processing, generative AI has also played a crucial role in addressing key issues in intelligent transportation systems (ITS), such as data sparsity, difficulty in observing abnormal scenarios, and in modeling data uncertainty. In this review, we systematically investigate the relevant literature on generative AI techniques in addressing key issues in different types of tasks in ITS tailored specifically for road transportation. 
First, we introduce the principles of different generative AI techniques. Then, we classify tasks in ITS into four types: traffic perception, traffic prediction, traffic simulation, and traffic decision-making. We systematically illustrate how generative AI techniques addresses key issues in these four different types of tasks. Finally, we summarize the challenges faced in applying generative AI to intelligent transportation systems, and discuss future research directions based on different application scenarios.
\end{abstract}




\maketitle

\section{Introduction}


The transportation system contains traffic infrastructure, equipment, and management strategies, including traffic control and traffic planning, all aimed at managing traffic flow and meet travel demands.
Its efficiency significantly affects societal and economic activities. Rapid urbanization has greatly increased the number of vehicles, leading to congestion and accidents. These challenges inconvenience travelers and disrupt urban operations, emphasizing the need of better traffic management. Recent advances in computing technology have enabled the development of Intelligent Transportation Systems (ITS). Traditionally, ITS covers air, rail, road, and marine transportation.  In this paper, we focus on urban road transportation and define ITS as an integrated system connecting people, roads, and vehicles.  

The foundational framework of ITS consists of four main components, including traffic perception, prediction, simulation, and decision-making. 
Traffic perception collects and analyzes large-scale data through sensors, cameras, and other monitoring devices to understand traffic dynamics, which in turn supports prediction models for forecasting future demands and conditions. Traffic simulation models traffic patterns under different scenarios, concentrating on realistic traffic state distributions to validate decision-making feasibility. The decision-making component leverage insights from perception, prediction, and simulation to implement effective traffic control measures and optimize transportation operations, while also generating new traffic data to evaluate their impact. Thus, these components form a cohesive closed-loop system that addresses the majority of ITS applications, enabling effective management, control, and optimization of transportation networks~\cite{lin2023generative,zhao2023autonomous}.

 Traditional deep learning methods have been widely adopted in ITS. These methods can process large number of traffic data, automatically recognize patterns, predict trends, and optimize decisions, thereby improving traffic efficiency and safety. For example, convolutional neural networks (CNNs) are used to analyze traffic camera images, identifying vehicles, pedestrians, traffic signs, and road conditions to monitor traffic flow and situations. Recurrent neural networks (RNNs) are effective for processing sequential data, such as time series data from traffic sensors, enabling predictions of traffic flow, congestion, and accident risk. Deep reinforcement learning (DRL) is applied to the field of autonomous driving, where models are trained to make effective driving decisions by simulating the behavior of intelligent agents, such as autonomous vehicles, in traffic environments. By utilizing these deep learning techniques, ITS can more effectively analyze road conditions, predict congestion, and provide adaptive decision support to optimize traffic efficiency.

However, with the increasing complexity of transportation networks, the integration of multiple transportation modes, and the growing demand for travel have made the traffic environment more complex. This complexity poses numerous challenges for traditional deep learning methods in addressing traffic-related issues.

\begin{itemize}
\item \emph{Large amounts of sparse or low-quality data}. Traffic data is often sparse and frequently suffers from missing values or noise contamination. Traditional deep learning methods impose high requirements on data quality, especially supervised learning approaches that typically require a substantial amount of labeled input and output data for supervised training. The scarcity or absence of traffic data can significantly impact model performance.

\item \emph{Rare abnormal scenarios}. Traffic environments exhibit high levels of dynamism, with occasional occurrences of anomalies such as traffic accidents that pose significant threats to traffic safety. These anomalies are challenging to capture as they are often rare events, resulting in a lack of relevant samples in training data. Traditional deep learning methods are more suitable for processing tasks under normal circumstances, and struggle to effectively address rare or previously unseen abnormal situations.

\item \emph{Unexplored modeling of uncertainty}.  In the transportation field, there are many factors that may cause uncertainty. For example, adverse weather conditions such as heavy rain, snow, or smog, as well as the uncertainty of various traffic participants' behaviors such as changing lanes and emergency braking, make traffic flow prediction more challenging. Traditional deep learning methods usually adopt deterministic approaches, relying on pre-processed features, thus making it difficult to effectively capture such uncertainty.

\end{itemize}

Generative Artificial Intelligence (Generative AI) technology has recently developed rapidly in recent years, enabling the automatic generation of content, including text, images, videos, and audio, based on user input or instructions. Several established techniques play key roles in generative AI. For example, Variational Auto-Encoder (VAE)~\cite{kingma2013auto} is a generative adversarial network that models and generates complex data. Generative Adversarial Network (GAN)~\cite{goodfellow2020generative} improves the quality of generated content through the adversarial process of generators and discriminators. Other techniques include normalizing flow~\cite{rezende2015variational} for generating complex data distributions, energy-based models~\cite{lecun2006tutorial} for enhancing generation quality, and generative models from physical processes~\cite{liu2023genphys} for generating content with physical constraints.  Additionally, diffusion probabilistic models~\cite{sohl2015deep} generate data by adding noise to data gradually to reduce its quality. Generative Pre-trained Transformer (GPT)~\cite{ghojogh2020attention} use pre-trained Transformer architectures\cite{vaswani2017attention} to generate text.

These generative AI techniques offer significant advantages in data generation, reconstruction, and data uncertainty modeling, which hold potential for addressing challenges in ITS~\cite{lv2023artificial}. 
For example, in autonomous driving scenarios, generative AI can generate high-fidelity driving scene images and videos, which are crucial for training and testing autonomous systems~\cite{krajewski2018data,kumar2023generative}. By using AI-generated simulation scenarios, autonomous vehicles can perform extensive driving simulations in virtual environments, improving their ability to make accurate real-world decisions. In traffic flow prediction tasks, generative AI technology can learn and model the distribution of traffic data, thus generating future traffic flow data~\cite{zhang2022stg,zand2023flow,wen2023diffstg}. This capability helps traffic authorities improve traffic planning and optimization.

\begin{table}[t!]
\scalebox{0.48}{
\begin{tabular}{|p{2.2cm}|l|l|l|}
\hline
                      & \textbf{Category}       & \textbf{Reference}                                     & \textbf{Description} \\ \hline
\centering \multirow{4}{*}[-1.2cm]{\textbf{ITS}}  & Comprehensive  & \begin{tabular}[c]{@{}l@{}}\cite{qureshi2013survey},~\cite{an2011survey},~\cite{singh2015recent},~\cite{ghosh2017intelligent},~\cite{figueiredo2001towards},~\cite{fayaz2018intelligent},~\cite{zear2016intelligent},~\cite{garg2023systematic},~\cite{xiong2012intelligent},~\cite{mandal2022framework},~\cite{tran2023factors},~\cite{rhoades2017survey},\cite{sharma2022introduction},~\cite{hassan2023intelligent},~\cite{ravi2022review}\end{tabular} & \begin{tabular}[c]{@{}l@{}}These works give a comprehensive review of ITS, covering \\ key technologies, infrastructure, and applications. However, \\ few generative AI techniques are systematically introduced.  \end{tabular}        \\ \cline{2-4} 
                      & Technologies   & \begin{tabular}[c]{@{}l@{}}DRL~\cite{haydari2020deep}, GNN~\cite{rahmani2023graph}, NLP~\cite{putri2021intelligent}, SVM~\cite{hao2020survey}, DL~\cite{veres2019deep,haghighat2020applications}, Positioning~\cite{du2021vulnerabilities,imparato2018vulnerabilities}, GAN~\cite{lin2023generative}, \\
                      Diffusion models~\cite{peng2024diffusion}, Blockchain~\cite{swain2006survey,das2023blockchain}, Edge computing~\cite{zhou2021intelligent,gong2023edge}, Dynamic pricing~\cite{saharan2020dynamic}, \\Fuzzy logic~\cite{swain2006survey},  Swarm intelligence~\cite{mathew2020swarm},  V2X downloading~\cite{wang2019survey}, Data mining~\cite{anand2018extensive,el2011data,zhu2018big}, \\Communication~\cite{deng2023review,camacho2018emerging,maimaris2016review,naja2013survey,shaaban2021visible,mare2016visible,khan2022dsrc} \end{tabular} &   \begin{tabular}[c]{@{}l@{}} These works conduct surveys on specific techniques within \\the field of in ITS.   \end{tabular}      \\ \cline{2-4} 
                      & Infrastructure & \begin{tabular}[c]{@{}l@{}}IoT~\cite{chand2018survey,patel2019survey,bhardwaj2019designing}, VANET~\cite{baras2018vanets,trivedi2019software}, Bluetooth~\cite{friesen2015bluetooth}, Radar-on-chip~\cite{saponara2019radar}, Hardware devices~\cite{damaj2022intelligent},\\ Distributed architecture~\cite{nasim2012distributed}, External infrastructure~\cite{cress2021intelligent} \end{tabular} & \begin{tabular}[c]{@{}l@{}} These works conduct surveys on specific components in ITS.   \end{tabular}          \\ \cline{2-4} 
                      & Applications   & \begin{tabular}[c]{@{}l@{}}Security management~\cite{ali2018issues,ben2015security,lamssaggad2021survey,harvey2020survey,amin2019big,affia2019security}, Traffic scheduling~\cite{nama2021machine}, Traffic detection~\cite{lin2022review},\\Driver behavior detection~\cite{chhabra2017survey}, Vehicle classification~\cite{gholamhosseinian2021vehicle,yang2018vehicle}, Metaverse~\cite{njoku2023prospects}, Urban mobility~\cite{mangiaracina2017comprehensive},  \\Urban monitoring~\cite{lima2017systematic}, Traffic sign detection~\cite{ellahyani2021traffic,luo2020traffic}, Traffic prediction~\cite{suhas2017comprehensive,yuan2021survey}, Traffic \\simulation~\cite{chen2024data}, Autonomous driving~\cite{altaf2021survey,cui2024survey},  Trust management~\cite{ma2011survey}, Spectrum regulation~\cite{choi2020survey} \end{tabular} & \begin{tabular}[c]{@{}l@{}} These surveys focus on specified applications in ITS.   \end{tabular}            \\ \hline
\centering \multirow{2}{*}[-0.3cm]{\textbf{\shortstack{Generative \\ AI}}} & Comprehensive  & \begin{tabular}[c]{@{}l@{}}\cite{cao2023comprehensive},~\cite{wu2023ai},~\cite{zhang2023complete},~\cite{foo2023aigc}\end{tabular} & \begin{tabular}[c]{@{}l@{}} These works conduct a comprehensive review of generative \\AI techniques.   \end{tabular}            \\ \cline{2-4} 
                      & Applications   & \begin{tabular}[c]{@{}l@{}}3D~\cite{li2023generative}, Biology~\cite{zhang2023survey}, ChatGPT~\cite{zhang2023one,wang2023survey}, Metaverse~\cite{qinem2023powering}, Edge cloud computing~\cite{wang2023overview},  Mobile\\ network~\cite{xu2023unleashing,du2023enabling}, Security \& Privacy~\cite{wang2023security,chen2023challenges}, Vehicle network~\cite{zhang2023generative} \end{tabular} & \begin{tabular}[c]{@{}l@{}} These surveys concentrate on the specific applications of \\generative AI techniques in a particular field.   \end{tabular}          \\ \hline
\end{tabular}}
\caption{Related reviews of ITS and generative AI. DRL: Deep reinforcement learning, GNN: Graph neural network, NLP: Natural language processing, V2X: Vehicle to everything. DL: Deep learning. IoT: Internet of things. VANET: Vehicle ad-hoc network. }\label{tab:surveys}
\vspace{-10mm}
\end{table}

\begin{figure}[t!]
	\centering
	\includegraphics[width=4.3in]{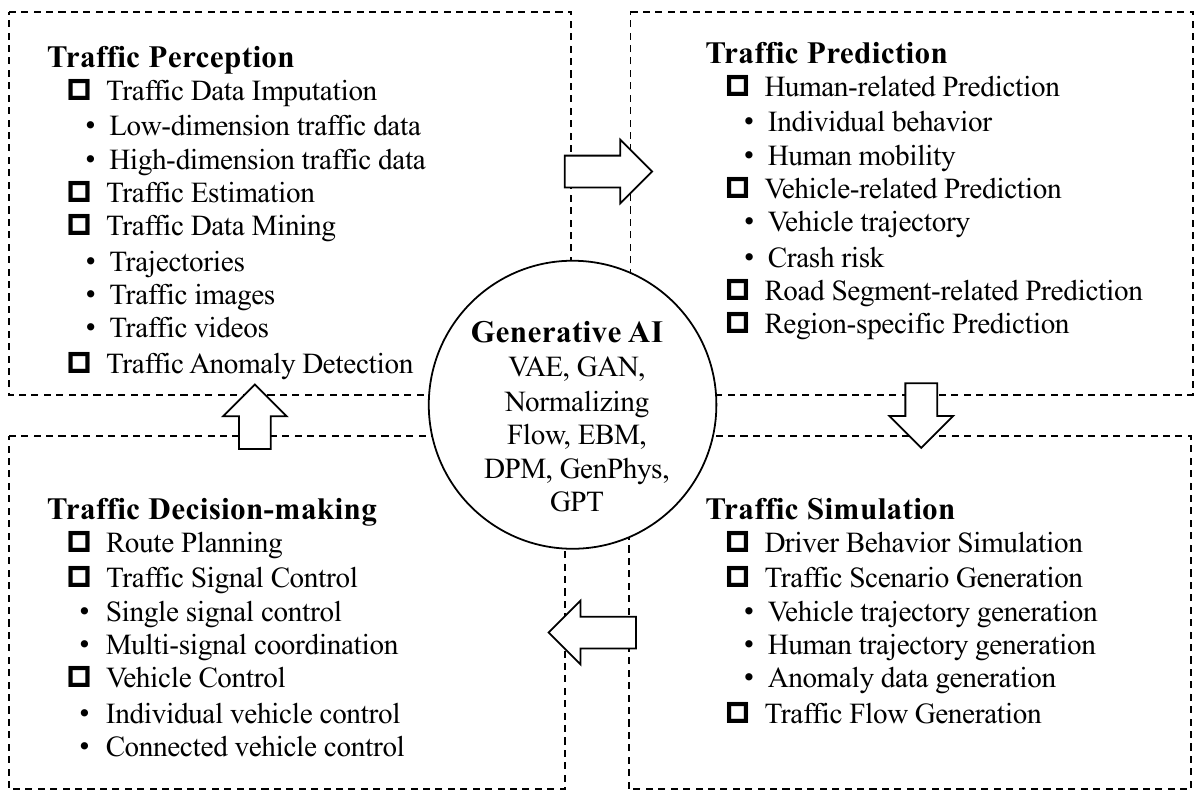}
 \vspace{-3mm}
	\caption{An overview of generative AI in intelligent transportation systems.}
	\label{fig:survey}
	\vspace{-7mm}
\end{figure}

Given the growing potential of generative AI in ITS, it is crucial to conduct an extensive review of past research that focuses on generative AI techniques applied to ITS.
 As shown in Table~\ref{tab:surveys}, numerous surveys have been conducted on ITS or generative AI. 
 However, most of these studies do not delve deeply into the practical applications of generative AI in ITS. 
 For example, while one study focuses on GANs~\cite{lin2023generative} and another on diffusion models~\cite{peng2024diffusion} in ITS, neither comprehensively addresses other generative AI techniques.
 Moreover, these surveys do not systematically examine how generative AI can overcome challenges faced by traditional deep learning approaches in ITS. 
 To bridge this gap, this paper explores the applications of generative AI techniques in traffic perception, prediction, simulation, and decision-making within ITS, as shown in Figure~\ref{fig:survey}. 
 Our goal is to provide a comprehensive analysis of their role in promoting intelligent transportation, and offer a constructive insight to help readers gain a deeper understanding of the challenges and opportunities of generative AI in intelligent transportation systems.  
 We expect this survey to provide useful insights for further research and advancement of intelligent transportation systems.

   Our survey focused on high-quality literature from reputable conferences, books, and journals. We referred to established archives like arXiv and Google Scholar, and major databases such as IEEE Xplore, ACM Digital Library, Springer, and ScienceDirect. Keywords like "Intelligent Transportation System" combined with generative AI terms (e.g., "variational autoencoder," "generative adversarial network," "GPT," "LLM") were used to identify relevant references. Additional filters included terms like "traffic signal," "traffic prediction," "traffic simulation," and "data imputation." We screened literature by examining titles, eliminating low-quality papers based on citation count, author affiliation, and publication authority, followed by an abstract review to assess contributions. This process ensured a foundation of credible, high-quality research.

 In conclusion, the primary contributions of this paper are illustrated as follows:
 \begin{itemize}
\item To the best of our knowledge, we are the first to offer an in-depth literature review of generative AI for intelligent transportation systems.
\item We provide a systematic introduction to mainstream generative AI techniques, conduct in-depth method comparisons from both horizontal and vertical perspectives, and provide a systematical analysis of how generative AI technology can effectively address key issues in intelligent transportation systems from the aspects of traffic perception, traffic prediction, traffic simulation, and traffic decision-making.
\item We discuss the open challenges encountered in applying generative AI technology in intelligent transportation systems, and explore potential directions for future research.
 \end{itemize}

  
   The structure of this paper is as follows. Section~\ref{sec:aigc} introduces mainstream generative AI techniques and gives a detailed comparative analysis. The following four sections respectively discuss the applications of generative AI technology in traffic perception, traffic prediction, traffic simulation, and traffic decision-making. Finally, in Section~\ref{sec:challenges}, we introduce the challenges faced by generative AI in the application of ITS, and look forward to future research directions.

\section{Fundamental Generative AI Technologies}\label{sec:aigc}

As described in previous works~\cite{cao2023comprehensive,foo2023aigc,wang2023survey,wu2023ai}, generative AI refers to a subset of AI algorithms that create personalized and high-quality content such as text, images, videos, and 3D assets based on user input or specific requirements. This section provides an overview of the fundamental technologies behind generative AI.

\subsection{Key Technologies}

\begin{figure}[t!]
	\centering
	\includegraphics[width=4.5in]{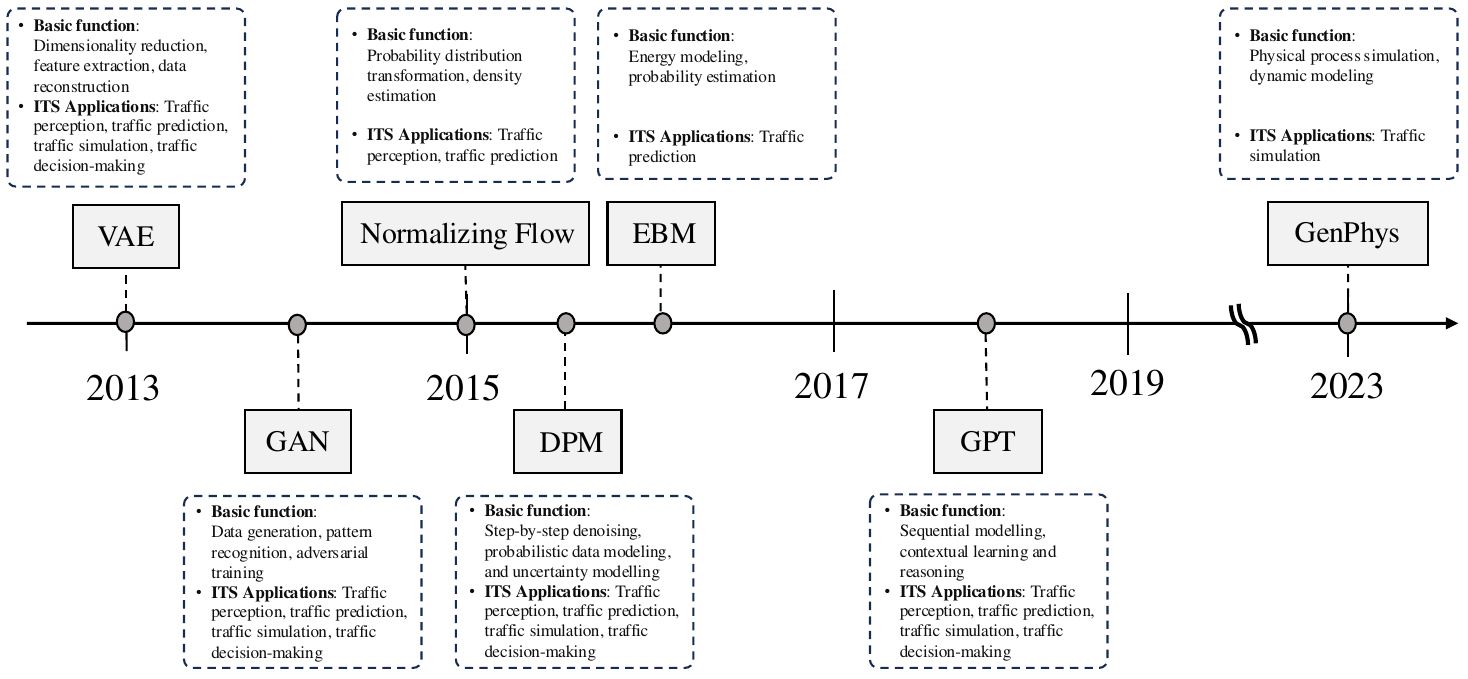}
 \vspace{-3mm}
	\caption{The history of generative AI techniques.}
	\label{fig:AIGC_history}
	\vspace{-6mm}
\end{figure}

With the rapid advancement of computer technology and computing power, generative AI has made significant progress over the past decade. 
In 2013, Kingma and Welling introduced VAE~\cite{kingma2013auto}, a probabilistic graphical model designed for data generation. In 2014, Ian Goodfellow et al. designed GANs~\cite{goodfellow2014generative}, which had a profound impact on the field of generative AI. 
In 2018, OpenAI’s GPT made significant contributions to the field. Beyond these, technologies like Normalizing Flow, Energy-Based Models (EBMs), Diffusion Probabilistic Models (DPMs) and Generative Models from Physical Process (GenPhys) have shown substantial potential for text and image generation.
Figure~\ref{fig:AIGC_history} depicts the evolution timeline of generative AI techniques, including VAE, GANs, Normalizing Flow, EBMs, DPMs, GenPhys, and GPT. In this subsection, we will provide an initial introduction to these techniques, followed by a comparative analysis.

\subsubsection{VAE}


Variational Autoencoders (VAEs)~\cite{kingma2013auto} aim to learn latent representations of data and generate new samples by mapping data to a latent space and using a decoder to reconstruct it. 

Given observed data $\mathbf{x}$ from an unknown distribution $p_{\theta}(\mathbf{x})$, VAEs introduce latent variables $\mathbf{z}$ and model $p_{\theta}(\mathbf{x}|\mathbf{z})$. The data distribution is represented as:
$p_{\theta}(\mathbf{x}) = \int_z p(z)p_{\theta}(\mathbf{x}|\mathbf{z}) \, dz.$
Since direct computation is intractable, VAEs use an encoder $q_{\phi}(\mathbf{z}|\mathbf{x})$ to approximate the posterior $p_\theta(\mathbf{z}|\mathbf{x})$, minimizing the Kullback-Leibler (KL) divergence between these distributions. 

In ITS applications, this approach allows capturing latent patterns in traffic data, generating synthetic traffic data that can be used for simulation, traffic prediction, and traffic decision-making.

\subsubsection{GAN}


Generative Adversarial Networks (GANs)~\cite{goodfellow2014generative} consist of a generator and a discriminator, trained together in an adversarial manner to create realistic data samples. The generator creates fake data from random noise, while the discriminator tries to distinguish it from real data
The goal is to reach a point where the generator produces data that the discriminator cannot reliably differentiate from real data, achieving a Nash equilibrium. This adversarial process enables ITS researchers to generate high-quality synthetic traffic scenarios that can be used for various applications, e.g., vehicle control. Generative Adversarial Imitation Learning (GAIL) extends GANs to learn policies by imitating expert behaviors without requiring explicit reward signals~\cite{ho2016generative}.

\subsubsection{Normalizing Flows}

Normalizing flows~\cite{rezende2015variational} transform a simple probability distribution into a complex one by applying a series of invertible transformations. These transformations are reversible, allowing for efficient calculation of probability densities.

Let $f$ be an invertible function that maps a random variable $\mathbf{z}$, sampled from a simple distribution $p(\mathbf{z})$, to a new variable $\mathbf{y} = f(\mathbf{z})$. The probability density of $\mathbf{y}$ is given by:
$
p\left(\mathbf{y}\right)=p(\mathbf{z})\left|\operatorname{det} \frac{\partial f^{-1}}{\partial \mathbf{y}}\right|=p(\mathbf{z})\left|\operatorname{det} \frac{\partial f}{\partial \mathbf{z}}\right|^{-1}.
$
By applying $K$ transformations to an initial variable $\mathbf{z}_0 \sim p_0(\mathbf{z}_0)$, we obtain a final variable $\mathbf{z}_K$ and its log probability density:
$
\ln p_K\left(\mathbf{z}_K\right)=\ln p_0\left(\mathbf{z}_0\right)-\sum_{k=1}^K \ln \left|\operatorname{det} \frac{\partial f_k}{\partial \mathbf{z}_{k-1}}\right|.
$

This allows starting from a simple distribution, like a Gaussian, and creating more complex, multi-modal distributions.
In the context of ITS, this allows modeling complex traffic data distributions while maintaining tractable probability densities. By applying normalizing flows to a simple distribution, we can accurately estimate the likelihood of specific traffic conditions, which is crucial for tasks like anomaly detection and probabilistic traffic forecasting.




\subsubsection{Energy-Based Models}
EBMs~\cite{lecun2006tutorial,du2019implicit} assign energy values to data points, with lower energy indicating a better fit. The goal is to assign lower energy to realistic points and higher energy to unrealistic ones, approximating the likelihood of observed data.

The probability density for data $\mathbf{x}$ is defined using a Boltzmann distribution:
$
p_\theta(\mathbf{x})=\frac{e^{-E_\theta(\boldsymbol{\mathbf{x}})}}{\int_{\tilde{\mathbf{x}} \in \mathcal{X}} e^{-E_\theta(\tilde{\mathbf{x}})}},
$
where $E_\theta(\mathbf{x})$ is an energy function parameterized by $\theta$.
To optimize EBMs, contrastive divergence compares positive samples (from real data) with negative samples (generated by the model), updating parameters to reduce energy for positives and increase energy for negatives. This is repeated iteratively to improve the model.
Due to their flexibility and robustness, EBMs are used to model complex traffic patterns and perform multivariate time series forecasting in ITS applications~\cite{yan2021scoregrad}.

\subsubsection{Diffusion Probabilistic Model}
DPMs~\cite{sohl2015deep} generate data by reversing a noisy process to recover the original data step by step.

In the forward process, a Markov chain successively adds Gaussian noise to the data $\mathbf{x}_0$, resulting in an approximate posterior:
$
q(\mathbf{x}_{1:T} \mid \mathbf{x}_0) = \prod_{t=1}^T q(\mathbf{x}_t \mid \mathbf{x}_{t-1}).
$

The reverse process starts from a noise distribution $p(\mathbf{x}_T) = \mathcal{N}(\mathbf{0}, \mathbf{I})$, and iteratively removes noise using learnable Gaussian transitions:
$
p_\theta(\mathbf{x}_{0:T}) = p(\mathbf{x}_T) \prod_{t=T}^1 p_\theta(\mathbf{x}_{t-1} \mid \mathbf{x}_t).
$

During training, the model predicts the noise added at each step, minimizing the difference between the predicted and actual noise.
In the ITS domain, DPMs are utilized to model and generate complex traffic patterns by progressively denoising noisy traffic data. This enables the creation of realistic synthetic traffic scenarios, which are valuable for traffic simulation, prediction and decision-making.

\subsubsection{Generative Models from Physical Process}
GenPhys~\cite{liu2023genphys} transforms physical partial differential equations (PDEs) into generative models, generalizing frameworks like diffusion models.

The goal is to generate new samples from observed data $p_{data}(\mathbf{x})$. A continuous physical process is described by:
$
\frac{d\mathbf{x}}{dt} = \mathbf{v}(\mathbf{x}, t),
$
which evolves the probability distribution $p(\mathbf{x}, t)$ as:
$
\frac{\partial p(\mathbf{x}, t)}{\partial t} + \nabla \cdot [p(\mathbf{x}, t) \mathbf{v}(\mathbf{x}, t)] - R(\mathbf{x}, t) = 0,
$
where $R(\mathbf{x}, t)$ represents particle birth or death. The goal is to design $p(\mathbf{x}, t)$, $\mathbf{v}(\mathbf{x}, t)$ such that $p(\mathbf{x}, t)$ matches $p_{data}(\mathbf{x})$ at $t=0$ and becomes independent as $T \to \infty$.

To generate samples, the reverse physical process is used to evolve $\mathbf{x}(T)$, drawn from $p(\mathbf{x}, T)$, backward to produce $\mathbf{x}(0)$. By incorporating GenPhys into ITS, we leverage the power of physics-informed generative modeling to improve traffic simulations, leading to better decision-making and more efficient transportation systems.

\subsubsection{Generative Pre-trained Transformer}

Generative Pre-trained Transformer (GPT) is a neural network designed for natural language processing (NLP) and text generation tasks. GPT uses a transformer decoder architecture~\cite{vaswani2017attention} with self-attention and feedforward layers to capture long-range dependencies in text. It processes inputs with positional encodings and word embeddings, and generates text in an autoregressive manner, token by token.
Pre-trained on large text datasets, GPT can be fine-tuned for specific tasks. 
These advanced capabilities can be leveraged to enhance various aspects of ITS, particularly in autonomous driving. For example, by fine-tuning GPT models on transportation-specific data, they can better understand complex driving scenarios, improving the decision-making processes of autonomous vehicles.

\subsubsection{Comparative Analysis of Generative AI Techniques}
In previous subsections, we introduced several major generative AI techniques, each with distinct advantages and limitations regarding performance, stability, flexibility, and computational requirements.
For example, VAEs effectively learn latent representations suitable for data with inherent uncertainty, with moderate computational demands and stable training, but may struggle to generate high-quality images.
GANs generate high-quality, diverse images using an adversarial process but face challenges like unstable training, mode collapse, and high computational demands. Table~\ref{tab:comparisonAIGC} summarizes these aspects for different generative AI techniques.

\begin{table}[t!]
\centering
\scalebox{0.54}{
\begin{tabular}{|p{3cm}|p{5cm}|p{5cm}|p{5cm}|p{5cm}|}
\hline
\textbf{Technology} & \textbf{Performance} & \textbf{Stability} & \textbf{Flexibility} & \textbf{Computational Requirements} \\ \hline

\centering \multirow{7}{*}{\textbf{\shortstack{Variational \\ Autoencoder~\cite{kingma2013auto}}}} & 
\begin{itemize}
    \item Learns effective latent representations
    \item Suitable for data with inherent uncertainty
    \item May produce blurry images
    
\end{itemize} & 
\begin{itemize}
    \item Relatively stable training process
\end{itemize} & 
\begin{itemize}
    \item Capable of probabilistic generation
\end{itemize} & 
\begin{itemize}
    \item Moderate computational resources needed
\end{itemize} \\ \hline

\centering \multirow{4}{*}{\textbf{\shortstack{Generative \\ Adversarial \\ Network~\cite{goodfellow2020generative}}}} & 
\begin{itemize}
    \item Generates high-quality and diverse images
\end{itemize} & 
\begin{itemize}
    \item Training can be unstable
    \item Prone to mode collapse
\end{itemize} & 
\begin{itemize}
    \item High flexibility in generation processes
\end{itemize} & 
\begin{itemize}
    \item Requires substantial computational power
\end{itemize} \\ \hline

\centering \multirow{5}{*}{\textbf{\shortstack{Normalizing \\ Flow~\cite{rezende2015variational}}}} & 
\begin{itemize}
    \item Accurately models complex probability distributions
\end{itemize} & 
\begin{itemize}
    \item Stable but computationally intensive
\end{itemize} & 
\begin{itemize}
    \item Supports both generation and inference
\end{itemize} & 
\begin{itemize}
    \item High computational and parameter requirements for high-dimensional data
\end{itemize} \\ \hline

\centering \multirow{5}{*}{\textbf{\shortstack{Energy-Based \\ Model~\cite{lecun2006tutorial}}}} & 
\begin{itemize}
    \item Models complex distributions effectively
    \item Robust to outliers
\end{itemize} & 
\begin{itemize}
    \item Training can be slow and requires careful tuning
\end{itemize} & 
\begin{itemize}
    \item Supports both generation and inference
\end{itemize} & 
\begin{itemize}
    \item High computational demands during training
\end{itemize} \\ \hline

\centering \multirow{5}{*}{\textbf{\shortstack{Diffusion\\ Probabilistic \\Model~\cite{sohl2015deep}}}} & 
\begin{itemize}
    \item Robust to noise and uncertainty
    \item Allows controlled generation
\end{itemize} & 
\begin{itemize}
    \item Stable training process
\end{itemize} & 
\begin{itemize}
    \item Flexible noise schedule
\end{itemize} & 
\begin{itemize}
    \item Longer generation time and complex training
\end{itemize} \\ \hline

\centering \multirow{4}{*}{\textbf{\shortstack{Generative Models \\from Physical\\ Process~\cite{liu2023genphys}}}} & 
\begin{itemize}
    \item Generates data grounded in physical processes
\end{itemize} & 
\begin{itemize}
    \item Stability depends on the physical model accuracy
\end{itemize} & 
\begin{itemize}
    \item Offers high interpretability
\end{itemize} & 
\begin{itemize}
    \item Requires prior knowledge of physical parameters
\end{itemize} \\ \hline

\centering \multirow{6}{*}{\textbf{\shortstack{Generative \\Pre-trained \\Transformer~\cite{ghojogh2020attention}}}} & 
\begin{itemize}
    \item Excels in natural language processing tasks
    \item Effective in modeling long-term dependencies
\end{itemize} & 
\begin{itemize}
    \item Generally stable once trained
\end{itemize} & 
\begin{itemize}
    \item Capable of generating continuous and coherent text
\end{itemize} & 
\begin{itemize}
    \item Requires substantial computing resources for pre-training
\end{itemize}
\\ \hline
\end{tabular}} 
\caption{Comparison of Generative AI technologies evaluated in terms of performance, stability, flexibility, and computational requirements. Partly referenced by the work~\cite{zhang2023generative}.} \label{tab:comparisonAIGC} \vspace{-8mm} 
\end{table}

\subsection{Important Generation Tasks}

\begin{figure}[t!]
	\centering
	\includegraphics[width=2.7in]{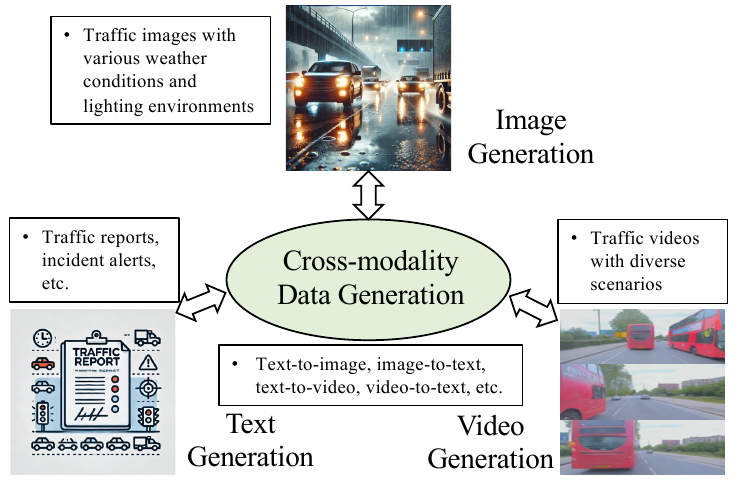}
 \vspace{-3mm}
	\caption{The relation between different generation tasks in ITS.}
	\label{fig:data_generation}
	\vspace{-6mm}
\end{figure}

The advancement of generative technologies has provided impressive solutions for various generative tasks. In this section, we will explore the applications of generative technologies in ITS from the perspectives of text, images, videos, and cross-modal data, as shown in Figure~\ref{fig:data_generation}.

\subsubsection{Text Generation}

Text generation focuses on using AI techniques to autonomously generate reasonable and meaningful text. In ITS, text generation is employed for tasks such as generating real-time traffic reports and automated incident alerts~\cite{grigorev2024incidentresponsegpt}. This capability provides better insight into traffic conditions and enables timely responses to dynamic traffic situations, thereby improving overall traffic management and safety.

\subsubsection{Image Generation}
Image generation involves using AI algorithms to create new images or visual content based on given information. In the context of ITS, image generation is utilized for generating synthetic images for testing the driving algorithms in autonomous vehicles~\cite{li2020style}. By generating realistic images reflecting various traffic scenarios, weather conditions, and lighting environments, generative methods help improve the robustness and performance of driving behavior decision-making models used in ITS.

\subsubsection{Video Generation}
Video generation involves creating a series of frames that are semantically coherent to form a video. This is particularly important in ITS for simulating traffic scenarios, training autonomous driving systems, and analyzing vehicle behaviors. The ability to generate realistic traffic videos allows researchers and engineers to test and validate ITS algorithms under a wide range of conditions without the need for extensive real-world data collection~\cite{zhao2024drivedreamer}.

\subsubsection{Cross-modality Data Generation}
Cross-modality data generation refers to generating data in one modality based on features from a different modality. In ITS, this is crucial for tasks such as converting textual descriptions of traffic situations into visual representations (text-to-image), generating textual reports from traffic camera images (image-to-text), or creating simulations of traffic scenarios based on textual inputs~\cite{ruan2024traffic}. These techniques enable more comprehensive traffic analysis and support the development of robust autonomous driving algorithms by providing diverse and rich datasets.

\section{Generative AI for Traffic Perception}

Traffic perception refers to the ability of ITS to collect and understand sensory information from the traffic environment. This information includes visual data from cameras, trajectory data from GPS-based devices, motion data from accelerometers, and environmental data such as weather and road conditions, as well as other traffic-related information like the presence of other vehicles, pedestrians, and road signs. Traffic perception is essential for making effective decisions and ensuring safety. In autonomous vehicles, traffic perception typically relies on cameras, LiDAR, millimeter-wave radar, and other technologies to collect and process data from the vehicle's surroundings, enabling autonomous driving and navigation. 

However, accurate perception and comprehension of the complex dynamics within traffic environments present several significant challenges. First, sensors may fail to provide data in certain situations, such as when cameras are obstructed. These data gaps can disrupt the continuous monitoring of traffic conditions, potentially leading to incomplete information. Second, sensor data collected in traffic environments is subject to various sources of noise, including sensor inaccuracies, weather conditions, and lighting variations. This noise introduces uncertainty, making it harder to derive precise information. Third, traffic environments are inherently complex and dynamic, involving numerous interacting factors such as vehicles, pedestrians, traffic signals, and road conditions, which accurately modeling and understanding traffic behavior challenging.

\begin{figure}[t!]
	\centering
	\includegraphics[width=3in]{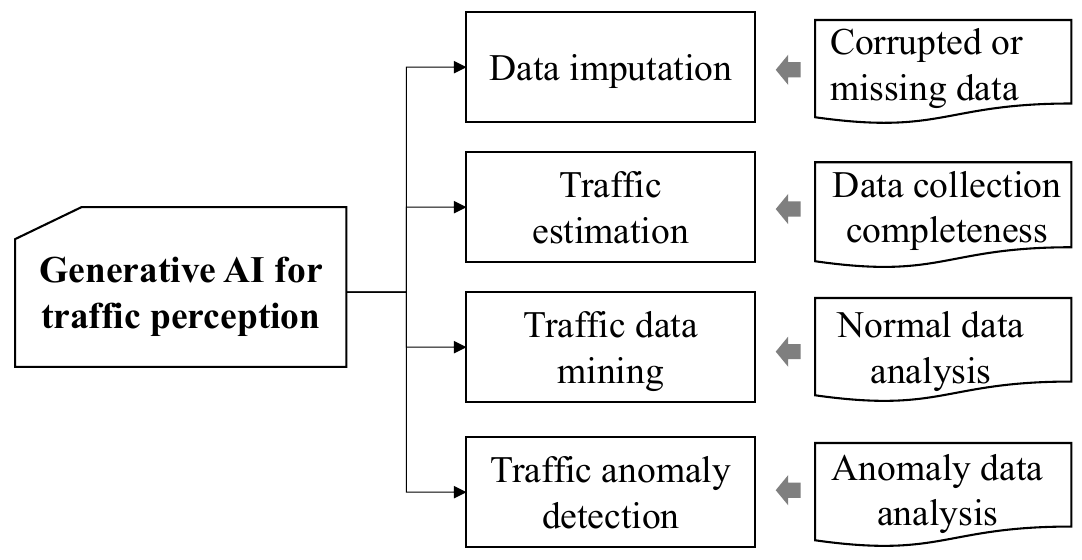}
 \vspace{-4mm}
	\caption{The relation between different topics in generative AI for traffic perception research.}
	\label{fig:traffic_perception}
	\vspace{-6mm}
\end{figure}

Recent advancements in generative AI technology offer promising solutions to these challenges. We carried an in-depth literature review focusing on the application of generative AI to address challenges in traffic perception. The reviewed literature covers various aspects of traffic perception, including data imputation, traffic estimation, data analysis and anomaly detection. Their relationships are illustrated in Figure~\ref{fig:traffic_perception}.

\subsection{Traffic Data Imputation}
Traffic data imputation involves learning a function, $f(\cdot)$, that takes corrupted or missing data $X$ as input to reconstruct the complete data $\hat{X}$. This can be expressed as $\hat{X}=f(X, \Psi)$, where $\Psi$ represents additional factors like the road network graph.

\textbf{Low-dimensional traffic data}. Low-dimensional traffic data, such as speed, flow, and travel time, is crucial for various applications like route planning and traffic prediction. However, missing data is common due to sensor failures or transmission issues.  To address this, researchers have employed advanced generative AI techniques to impute these missing data.

\emph{VAE}. Boquet et al.\cite{boquet2019missing} developed an online unsupervised VAE-based method for imputing missing traffic data, enhancing forecasting accuracy while enabling data augmentation, compression, and classification. Building on this, Chen et al.\cite{chen2021learning} introduced the STVAE model, which incorporates 3D gated convolutions, a multi-attention network, and Sylvester normalizing flows to better capture complex spatio-temporal dependencies. 

\emph{GAN}. Recently, more researchers have adopted GAN-based methods to address this task~\cite{chen2019traffic,han2020content,kazemi2021igani}. 
For example, 
Chen et al.~\cite{chen2019traffic} proposed using real or corrupted data as latent codes and introducing representation loss within a GAN model.
IGANI~\cite{kazemi2021igani} introduced a novel iterative GAN imputation architecture, which iterated
over imputed data while preserving the reversibility of the generative imputer.
In particular, numerous works proposed to incorporate the spatio-temporal modeling in GAN-based architectures~\cite{huang2021deep,yang2021st,zhang2021missing,zhang2021gated,zhang2021gated,yang2022st,huang2022data,shen2022traffic,li2023dynamic,li2023multistate,peng2023generative}.
For example, the authors in~\cite{li20183d} introduced a 3D convolutional GAN that excels in capturing intricate traffic dynamics through fractionally strided and standard 3D CNNs for the generator and discriminator, respectively.
STGAN~\cite{yuan2022stgan} enhanced GANs with generation and center losses, along with skip connections and dilated convolutions, achieving superior imputation while maintaining data consistency. 
GAE-GAN-LSTM~\cite{xu2022generative} integrated a graph autoencoder with GANs and LSTM, offering robust imputation across various loss ratios by effectively modeling complex traffic patterns. These innovations highlight the superior capability of advanced GAN-based methods in capturing and reconstructing the complex spatio-temporal relationships in traffic data, offering more reliable and versatile solutions.

\emph{DPM}. 
Recently, diffusion models have attracted more researchers' attentions in traffic data imputation~\cite{liu2023pristi,zhang2023sasdim,yun2023imputation}.
For instance, Liu et al.~\cite{liu2023pristi} introduced PriSTI, an enhanced conditional diffusion framework that handles diverse missing patterns and high missing rates by integrating conditional feature extraction and advanced noise estimation. This method mitigates error accumulation seen in autoregressive models and incorporates geographic relationships, offering a more context-aware imputation.
Similarly, SaSDim~\cite{zhang2023sasdim} enhanced diffusion models with a self-adaptive noise scaling mechanism and a spatial-temporal global convolution module. Their novel loss function normalizes noise intensity, reducing the negative impact of noise and capturing dynamic dependencies more effectively. These methods illustrate the potential of diffusion models to address spatio-temporal imputation tasks by leveraging their inherent flexibility in handling noise and capturing complex data dependencies. 

\begin{table}[]
\scalebox{0.7}{
\begin{tabular}{|l|l|l|}
\hline
\textbf{Method}               & \textbf{Spatio-Temporal Modeling} & \textbf{Papers} \\ \hline
\multirow{2}{*}{\textbf{VAE}} &         MLP        &     \cite{boquet2019missing}   \\ \cline{2-3} 
                     &     3D gated CNN, multi-attention mechanism        &     \cite{chen2021learning}   \\ \hline
\multirow{13}{*}{\textbf{GAN}} &     3D CNN            &   \cite{li20183d}     \\ \cline{2-3} 
                    &     Dilated CNN            &   \cite{yuan2022stgan}     \\ \cline{2-3}
                    &       CNN        &    \cite{yang2022st,zhang2021generative}    \\ \cline{2-3}
                    &  CNN, self-attention mechanism& \cite{zhang2021missing}        \\ \cline{2-3}
                    &       Temporal: Gramian Angular Summation Field          &    \cite{huang2021deep}    \\ \cline{2-3}
                     &       Temporal: discrete Wavelet transform          &    \cite{huang2022data}    \\ \cline{2-3}
                      &       Temporal: multi-view temporal factorizations          &    \cite{li2023dynamic}    \\ \cline{2-3}
                     &     \begin{tabular}[c]{@{}l@{}}  Spatial: learnable bidirectional attention map; Temporal: multi-channel matrix    \end{tabular}     &    \cite{yang2021st}    \\ \cline{2-3}
                     &     \begin{tabular}[c]{@{}l@{}}  Spatial: GCN; Temporal: LSTM    \end{tabular}     &    \cite{xu2022generative,shen2022traffic}    \\ \cline{2-3}
                     &     \begin{tabular}[c]{@{}l@{}}  Spatial: multi-layer FCN; Temporal: LSTM    \end{tabular}     &    \cite{li2023multistate}    \\ \cline{2-3}
                     &     \begin{tabular}[c]{@{}l@{}}  Spatial: dynamic GCN; Temporal: multi-head self-attention network   \end{tabular}     &    \cite{peng2023generative}    \\ \cline{2-3}
                     &    \begin{tabular}[c]{@{}l@{}}  Spatial: spatial attention GCN; Temporal: self-attention GRU     \end{tabular}          &    \cite{zhang2021gated}  \\ \hline
\multirow{3}{*}{\textbf{DPM}} &       Spatial: GNN          &    \cite{yun2023imputation}    \\ \cline{2-3}
                    &      \begin{tabular}[c]{@{}l@{}}  Spatial: GCN; Temporal: Transformer     \end{tabular}           &   \cite{liu2023pristi}     \\ \cline{2-3} 
                     &    \begin{tabular}[c]{@{}l@{}}  Spatial: dynamic GCN; Temporal: global temporal convolution     \end{tabular}             &   \cite{zhang2023sasdim}     \\ \hline
\end{tabular}}
\caption{Spatio-temporal modeling in generative AI methods for low-dimensional traffic data imputation.}\label{tab:data_imputation}
\vspace{-11mm}
\end{table}

Finally, to provide a clear illustration of the spatio-temporal modeling within the generative AI framework for this task, we summarize the related works in Table~\ref{tab:data_imputation}.

\textbf{High-dimensional traffic data}. High-dimensional traffic data, such as traffic videos and 3D LiDAR vision, plays a crucial role in enhancing situational awareness and decision-making, particularly in autonomous driving. However, the loss of high-dimensional data remains a common challenge in real-world applications. To address this, Wu et al.~\cite{wu2022traffic, wu2021high} utilized feature pyramid networks within GAN frameworks to extract multi-scale semantic features from traffic videos. This multi-scale approach enhanced the integration of spatial and temporal information, ensuring continuity and accuracy in the imputed frames. The incorporation of local and dual-branch discriminators further guaranteed spatio-temporal consistency, which was crucial for applications requiring precise motion and context understanding, such as traffic monitoring and autonomous navigation.
Tu et al.~\cite{tu2021incomplete} extended GAN applications to 3D LiDAR point clouds with their Point Fractal Network (PF-Net). By repairing incomplete point clouds, PF-Net restores vital geometric and semantic information, thereby improving object detection and collision avoidance in autonomous vehicles. This GAN-based repair not only enhances data integrity but also supports the real-time requirements of autonomous driving systems.

\subsection{Traffic Estimation}
Traffic estimation aims to learn a estimator to estimate the current traffic condition of all roads in the transportation network. This task differs from traffic data imputation in that it addresses the challenge of incomplete data collection due to constraints such as cost and time limitations.

Several works designed GAN-based approaches to estimate traffic condition~\cite{liang2018deep,yu2019real,zhang2021c,mo2022quantifying,tian2022pattern,zhang2022strans,xu2023diffusion}. 
GAA~\cite{liang2018deep} merged traffic-flow theory with GANs using LSTM networks to effectively capture spatio-temporal correlations. This integration not only improved estimation accuracy compared to Bayesian methods but also demonstrated the potential of combining theoretical models with deep learning.
GCGA~\cite{yu2019real} combined GCN with GANs, capturing the spatial dependencies and leading to more accurate traffic condition estimations.
Innovations like Curb-GAN~\cite{zhang2020curb} and GE-GAN~\cite{xu2020ge} further refined these models by incorporating dynamic convolutional layers and graph embeddings, enabling more precise and context-aware traffic estimations.
PA-GAN~\cite{tian2022pattern} integrated Bayesian inference to adaptively adjust parameters based on contextual features, effectively handling multi-modal traffic patterns and reducing estimation biases.
Additionally, Zhang et al.~\cite{zhang2021c} models the complex correlations between road traffic and urban conditions in their proposed conditional GAN-based model, further improving estimation accuracy. 

Unlike them, the authors in~\cite{yuan2023traffic} leveraged denoising DPM by parameterizing noise factors, transforming traffic matrix estimation into a gradient-descent optimization task. This method demonstrated superior performance in both traffic matrix synthesis and estimation. 
Lei et al.\cite{lei2024conditional} further extended diffusion models with their Conditional Diffusion Framework with Spatio-Temporal Estimator (CDSTE), which integrates probabilistic diffusion with spatio-temporal networks. CDSTE excels in inferring complete traffic states at sensor-free locations, providing reliable probabilistic estimates that support robust traffic management and control decisions.

\subsection{Traffic Data Mining}
In this section, we introduce the related works that utilize generative AI techniques to analyze traffic data from various sources, including trajectories, traffic images, and traffic videos, as summarized in Table~\ref{tab:traffic_analysis}.

\textbf{Trajectories}. Trajectories are sequences of timestamped positions that describe the movement of an object over time, providing valuable insights for analyzing various aspects such as urban transportation modes and human decision-making preferences. To identify the transportation modes, the authors in~\cite{zhang2022geosdva} introduced a semi-supervised Dirichlet VAE that integrates geographic data with motion features. This model effectively leveraged limited labeled GPS data and accounts for geographic factors, achieving superior classification accuracy on real-world datasets. Similarly, TrajGAIL~\cite{zhang2020trajgail} used Variable Length Markov Decision Processes and deep neural networks to capture long-term dependencies in human decisions. This model outperformed traditional Markov-based models, demonstrating enhanced accuracy in replicating complex behaviors.

\textbf{Traffic images}. In autonomous driving scenarios, traffic images captured by cameras provide essential information about the surrounding environment, allowing autonomous vehicles to make effective decisions.
To overcome the challenge of limited image diversity in real urban environments, Li et al.~\cite{li2020style} introduced a GAN-based framework that performs diverse urban image style transfers, such as day-to-night transitions, while effectively preserving foreground objects and structural details. This method outperformed baseline models by eliminating distortions and maintaining high-quality transformations, thereby increasing the variability and reliability of training data for autonomous systems.
Zhao et al.~\cite{zhao2022raindrop} addressed image degradation from raindrops on in-vehicle cameras with a GAN-based removal network that integrates visual attention and content perception. This approach not only enhances image quality but also improves traffic object detection accuracy.

Unlike them, some researchers focus on extracting valuable information from traffic images for tasks such as vehicle detection~\cite{wang2021small}, license plate identification~\cite{lee2018accurate,han2020license,ibrahim2022license,boby2023iterative}, traffic density recognition~\cite{gawali2023dual}, driving behaviors~\cite{wijaya2022deepdrive} and map enrichment~\cite{zhong2022probabilistic}. These tasks often face challenges related to low resolution and inadequate lighting conditions.
For example, the authors in~\cite{hassan2023unleashing} enhanced vehicle detection in low-light conditions by integrating Pix2PixGAN and CycleGAN with a YOLOv4-based algorithm. This method effectively overcomes the limitations of traditional vision techniques in poor lighting, demonstrating substantial performance improvements that enhance autonomous vehicle reliability.
In the domain of license plate detection, Boby et al.~\cite{boby2023iterative} compared a GAN-based super-resolution model with diffusion-based iterative refinement models. This comparison highlighted the importance of selecting the appropriate super-resolution technique based on specific application requirements, with GANs offering significant advantages in measurable performance metrics.
In addition, the authors in~\cite{gawali2023dual} presented a dual-discriminator conditional GAN for recognizing traffic density in both homogeneous and heterogeneous traffic scenarios. The dual-discriminator architecture, optimized using Giza pyramids construction, enhanced the model’s ability to distinguish between varying traffic densities with improved accuracy and reduced loss. 

\begin{table}[t!]
\scalebox{0.8}{
\begin{tabular}{|c|l|l|l|}
\hline
\multicolumn{1}{|l|}{\textbf{Category}}  & \textbf{Task}                                    & \textbf{Method}           & \textbf{Papers}                                                       \\ \hline
\multirow{2}{*}{\textbf{Trajectory}}     & Transportation mode identification      & VAE              & \cite{zhang2022geosdva}                     \\ \cline{2-4} 
                                & Decision-making   behavior analysis     & GAN              & \cite{zhang2020trajgail}                    \\ \hline
\multirow{10}{*}{\textbf{Traffic image}} & Image transformation                    & GAN              & \cite{li2020style}                          \\ \cline{2-4} 
                                & Image denoise                           & GAN              & \cite{zhao2022raindrop}                     \\ \cline{2-4} 
                                & Image resolution enhancement            & GAN              & \cite{ibrahim2022license,boby2023iterative} \\ \cline{2-4} 
                                & Vehicle classification                  & GAN              & \cite{wang2021small}                        \\ \cline{2-4} 
                                & Vehicle detection                       & GAN              & \cite{hassan2023unleashing}                 \\ \cline{2-4} 
                                & License plate recognition               & GAN              & \cite{lee2018accurate,han2020license}       \\ \cline{2-4} 
                                & Traffic density identification          & GAN              & \cite{gawali2023dual}                       \\ \cline{2-4} 
                                & Behavior detection                      & GAN              & \cite{wijaya2022deepdrive}                  \\ \cline{2-4} 
                                & Road annotation                         & GPT              & \cite{juhasz2023chatgpt}                    \\ \cline{2-4} 
                                & Flow inference                          & Normalizing Flow & \cite{zhong2022probabilistic}               \\ \hline
\multirow{5}{*}{\textbf{Traffic video}}  & Vehicle-road user interaction detection & VAE              & \cite{cheng2021interaction}                 \\ \cline{2-4} 
                                & Object segmentation                     & GAN              & \cite{patil2020end}                         \\ \cline{2-4} 
                                & Activity classification                 & GAN              & \cite{krishna2014combination}               \\ \cline{2-4}
                                & BEV perception                 & DPM              & \cite{zou2024diffbev}
                                \\ \cline{2-4}
                                & Accident video understanding                  & DPM              & \cite{fang2024abductive}
                                \\ \hline
\end{tabular}}
\caption{Related works about traffic data mining.}\label{tab:traffic_analysis}
\vspace{-10mm}
\end{table}

\textbf{Traffic videos}. Compared to traffic images, traffic videos provide a wealth of redundant traffic information for autonomous vehicles, guaranteeing both safety and efficiency in navigation. A pivotal stage in numerous intelligent video processing applications, such as vehicle behavior analysis, is the segmentation of moving objects.
RMS-GAN~\cite{patil2020end} is an end-to-end GAN using a recurrent technique, which integrates foreground probability information using residual and weight-sharing techniques to achieve precise segmentation.
Krishna et al.~\cite{krishna2014combination} progressively combined generative and discriminative models to efficiently classify activities in traffic videos.
Cheng et al.~\cite{cheng2021interaction} proposed a deep conditional generative model to detect interactions between vehicles and vulnerable road users. This model utilized a conditional VAE-based framework with Gaussian latent variables to capture the behavior of road users and predict interactions in a probabilistic manner.
DiffBEV~\cite{zou2024diffbev} addressed the issue of noisy Bird’s Eye View (BEV) representations in autonomous driving by using a conditional diffusion model to improve BEV feature quality through progressive refinement.
The authors in~\cite{fang2024abductive} proposed an abductive framework AdVersa-SD for accident video comprehension. AdVersa-SD employs an object-centric video diffusion approach to learn cause-effect relationships of accidents, thereby enhancing safe driving perception.

\subsection{Traffic Anomaly Detection}
Traffic anomalies refer to unusual events, patterns, or behaviors within a transportation system. These anomalies can disrupt the regular traffic flow or pose potential safety risks. Numerous studies have explored the use of generative AI techniques for traffic anomaly detection~\cite{qiu2019driving,lin2020automated,kumar2020hybrid,xie2023novel}.
Chen et al.~\cite{chen2021multi} presented a multi-modal GAN model to detect and classify the traffic events, leveraging data from multiple modalities to enhance detection accuracy.
The authors in~\cite{santhosh2021vehicular} proposed a hybrid CNN-VAE architecture that employs color gradient representations and semi-supervised annotation, achieving a 1-6\% increase in classification accuracy and a 30-35\% boost in anomaly detection through effective feature extraction and semi-supervised labeling.
The work by Kang et al.~\cite{kang2022traffic} addressed congestion anomaly detection by integrating conditional normalizing flows with kernel density estimators to effectively capture spatial-temporal dependencies in multivariate time series data. Their framework not only demonstrated superior performance, but also ensured robustness across diverse traffic scenarios. 
A-VAE~\cite{aslam2023vae} integrated 2D CNN and BiLSTM layers with an attention mechanism for enhanced representation learning in traffic video anomaly detection. This approach demonstrated strong real-time processing capabilities, highlighting its practical applicability for live traffic monitoring systems.
Li et al.~\cite{li2024difftad} presented DiffTAD, a diffusion model-based framework for vehicle trajectory anomaly detection that addresses training instability and mode collapse issues prevalent in GANs and VAEs. By modeling anomaly detection as a noisy-to-normal process and incorporating Transformer-based encoders for temporal and spatial dependencies, DiffTAD achieved superior performance and robustness in diverse traffic environments.
Lastly, Yan et al.~\cite{yan2023feature} introduced a novel diffusion model–based method for video anomaly detection that utilizes two denoising diffusion implicit modules to predict and refine the motion and appearance features of video frames. The proposed method effectively learned the distribution of normal samples, enabling accurate detection of anomalies.

Cyberattacks would pose significant safety risks in the context of automated driving. 
To address it, researchers have proposed a variety of solutions leveraging generative models. 
Zhao et al.~\cite{zhao2022gvids} presented GVIDS, a GAN-based intrusion detection system for connected vehicles that distinguishes legitimate CAN messages from malicious ones. Although GVIDS is suitable for dynamic vehicular environments, its reliance on specific CAN message characteristics may limit adaptability to new attack types.
Ding et al.~\cite{ding2019trojan} exposed the vulnerability of deep generative models in autonomous driving to Trojan attacks through data poisoning. Their work underscored the critical risk of maliciously implanted behaviors that activate under specific triggers, highlighting the need for robust data integrity measures. 
Li et al.~\cite{li2022detecting} developed a GAN-based real-time anomaly detection model capable of identifying subtle adversarial manipulations in vehicle and sensor data, demonstrating its potential for dynamic environments.
In addition, the authors in~\cite{devika2024vadgan} presented an unsupervised GAN-based framework optimized with LSTM networks to analyze vehicle dynamics, effectively detecting eleven distinct attack types. By integrating LSTM architectures, the framework effectively handled extensive vehicular data, demonstrating robustness in maintaining the integrity of connected and autonomous vehicle networks.

\subsection{Discussion}

We have summarized the key insights and challenges of generative AI in traffic perception. Generative AI techniques such as VAE, GAN, and DPM have shown advantages in handling missing low- and high-dimensional traffic data, particularly in capturing complex spatiotemporal dependencies, improving traffic prediction and management. However, their ability to handle data noise and dynamic changes varies, requiring scenario-specific selection. Generative AI has also enhanced traffic estimation by integrating traditional traffic theory and deep learning, significantly improving accuracy. In traffic data mining, progress has been made in trajectory analysis, image processing, and video analysis, but challenges related to data diversity and model generalizability exist. In anomaly detection, generative AI effectively identifies both regular and anomalous patterns, but real-time performance and adaptability need improvement. Future research should focus on optimizing spatiotemporal modeling, improving data robustness, and integrating integration with traditional traffic theory to optimize ITS.

\section{Generative AI for Traffic Prediction}\label{sec:prediction}
\begin{figure}[t!]
	\centering
	\includegraphics[width=3.5in]{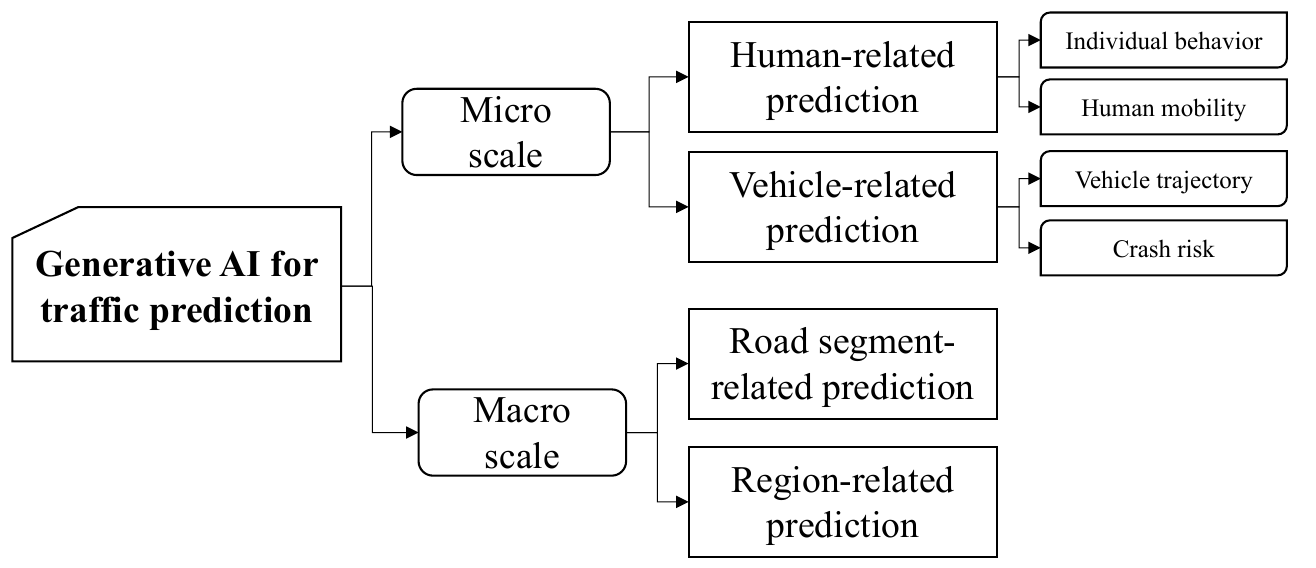}
 \vspace{-3mm}
	\caption{The relation between different topics in generative AI for traffic prediction research.}
	\label{fig:traffic_prediction}
	\vspace{-6mm}
\end{figure}

Traffic prediction refers to estimate or forecast future traffic conditions, encompassing various aspects such as travel demand, travel time, traffic flow, and the movements and behaviors of both vehicles and individuals within the transportation system. Traffic prediction is vital for urban traffic planning and management. However, achieving accurate predictions presents significant challenges. First, urban traffic systems exhibit complex spatial and temporal dynamics. Second, traffic conditions are highly dynamic, susceptible to rapid changes due to factors like real-time incidents, ongoing construction activities, and even human behavior. 
Third, obtaining high-quality data from various sensors, cameras, and GPS devices is important for traffic prediction, but such data is often limited and incomplete. Generative AI technology offers powerful solutions to overcome these challenges. In our comprehensive literature review, we concentrate on the utilization of generative AI in addressing traffic prediction challenges. We categorize these works into four main aspects: human-related, vehicle-related, road segment-related and region-related traffic prediction, encompassing the key components of transportation systems. Their relation is given in Figure~\ref{fig:traffic_prediction}.

\subsection{Human-related Traffic Prediction}
Human-related traffic prediction involves forecasting individual behavior and human mobility.

\textbf{Individual behavior}. We discuss individual behavior prediction based on their roles within transportation systems. For drivers, modeling and predicting driver behavior plays a crucial role in enhancing safety and preventing risky actions~\cite{ivanovic2020multimodal,bao2021prediction}. For example, Bao et al.~\cite{bao2021prediction} introduced a probabilistic sequence-to-sequence approach utilizing a conditional VAE to predict various driving behaviors. 
For pedestrians, accurately forecasting human motion behavior is vital for autonomous vehicles to plan preemptive actions and avoid collisions, ensuring pedestrian safety~\cite{sun2019intent,cui2021efficient}. 
For instance, 
in~\cite{cui2021efficient}, the authors proposed a temporal convolutional GAN for high-fidelity future pose prediction, incorporating two discriminators: one for fidelity assessment and another for consistency evaluation in long sequences.
UTD-PTP~\cite{tang2024utilizing} introduces a universal transformer diffusion modeling framework that combines transformer architectures with diffusion-based uncertainty modeling to accurately predict pedestrian trajectories in complex autonomous driving environments.

\textbf{Human mobility}. Human mobility refers to an individual's movement from one location to another. Feng et al.~\cite{feng2023ilroute} presented a graph-based imitation learning framework that adeptly models courier decision-making in route selection. By utilizing a multi-graph neural network, ILRoute effectively captured the complex, multi-source features influencing route choices, while incorporating mobility regularity and personalized constraints to navigate the expansive decision-making space. This approach improved prediction accuracy across multiple real-world datasets and online evaluations. Building on the necessity to understand and predict human mobility patterns during unprecedented events like the pandemic, Bao et al.~\cite{bao2020covid} introduced a spatio-temporal conditional GAN Covid-GAN to estimate mobility changes under varying public health conditions and policy interventions. Further, the authors extended their initial model to Covid-GAN+~\cite{bao2022covid} addressed spatial heterogeneity by using the local Moran statistic and outliers by redesigning the training objective to learn the estimated mobility changes. This method achieved a more accurate and robust estimation of human mobility.

\subsection{Vehicle-related Traffic Prediction}
In the reviewed literature, vehicle-related traffic prediction tasks encompass vehicle trajectory prediction and crash risk prediction.

\textbf{Vehicle trajectory}. 
A vehicle trajectory refers to the path a vehicle takes from one location to another over a specific time period. Precise vehicle trajectory prediction is vital for a variety of traffic-related applications, including traffic management, navigation, and route planning.

Most problems of vehicle trajectory prediction are defined as using historical trajectories to infer future trajectories. Formally, given a vehicle $i$ with a historical trajectory $\mathbf{x}_i=\{r_i^t\}$ for $t=1, 2, \dots, T$, where $r_i^t$ represents the location at time $t$, the goal is to learn a mapping function $g(\cdot)$ to predict the future trajectory $\mathbf{y}_i=\{r_i^t\}$ for $t=T+1, \dots, T+n$.

However, achieving precise trajectory predictions is a challenging task for several reasons. First, vehicle movements are influenced by various factors, including driver behavior, traffic conditions, road infrastructure, and unexpected events. Second, traffic conditions can change rapidly due to incidents, congestion, or accidents. Third, the accuracy of trajectory prediction models heavily relies on the quality and quantity of input data, such as GPS data. Noisy or incomplete data can lead to inaccurate predictions. To address these challenges, more researchers proposed different methods based on generative AI technology to predict vehicle trajectory.



\emph{VAE}. Several researchers have designed prediction models based on VAE~\cite{cheng2020context,kim2021driving,xing2022multi,oh2022cvae}. 
DVAE~\cite{neumeier2021variational} integrated expert knowledge into the decoder component of the VAE, which can provide the similar prediction accuracy and enhance interpretability.
The authors in~\cite{jagadish2021autonomous,jagadish2022conditional} utilized conditional VAEs to encode past trajectories of vehicles and traffic scenes for vehicle path prediction. By conditioning the network on these inputs, their approach effectively captured the multimodal nature of vehicle movements in various traffic scenarios.
Gui et al.~\cite{gui2022visual} incorporated additional dynamic information like yaw angle and velocity in their proposed conditional VAE framework. This approach allows for a more comprehensive understanding of vehicle interactions and movements, resulting in superior performance on both established and newly introduced simulated datasets.

\emph{GAN}. GAN is also an alternative technique for trajectory prediction~\cite{roy2019vehicle,li2021vehicle,zhou2021sa}. 
For example, Kang et al.~\cite{kang2020vehicle} applied social GAN, originally designed for pedestrian trajectory prediction, for the prediction of vehicle trajectories. While this approach demonstrated better prediction accuracy, its reliance on models initially tailored for pedestrian dynamics may limit its adaptability to the distinct behaviors exhibited by vehicles.
Differently, Zhou et al.~\cite{zhou2021sa} introduced the self-attention social GAN to handle long trajectory sequences and complex vehicle interactions. The self-attention mechanism helped capture correlations in trajectory data, reducing the loss of important information over time. However, the model’s complexity may limit its suitability for real-time applications requiring high computational efficiency.
KI-GAN~\cite{wei2024ki} was a knowledge-informed GAN that integrated traffic signal data and multi-vehicle interactions through a specialized attention pooling mechanism. This model effectively incorporated orientation and proximity information for accurate predictions but relied heavily on the quality of input data, which can vary across urban environments.


\emph{GPT}. Recently, GPT models has great potential in vehicle trajectory prediction~\cite{su2022crossmodal,feng2023trtr,vishnu2023improving}. Su et al.~\cite{su2022crossmodal} proposed a crossmodal transformer based generative framework that integrates multiple modalities and pedestrian attributes to predict pedestrian trajectories more accurately.
Similarly, Feng et al.~\cite{feng2023trtr} developed a pre-trained large traffic model using the Transformer architecture. This model has ability to capture the diverse trajectories within a population of vehicles,  leading to higher accuracy in vehicle trajectory prediction.
However, the high computational demands of such large model may pose challenges for real-time deployment in autonomous systems. 

\emph{DPM}. Diffusion models also attract more researchers' attentions. For instance, EquiDiff~\cite{chen2023equidiff} combined the conditional diffusion model with an equivariant transformer, leveraging the geometric properties of position coordinates and social interactions among vehicles for accurate prediction.

\textbf{Crash risk}.
Crash risk prediction is important in traffic safety management and precaution. Due to imbalanced data between crash and non-crash cases, accurately predicting crash risk is challenging. To deal with this, Cai et al.~\cite{cai2020real} proposed a deep convolutional GAN model to generate synthetic crash data. This approach effectively preserves the heterogeneity of non-crash scenarios, enabling the use of the entire non-crash dataset for training predictive models. Similarly, in~\cite{man2022wasserstein}, Wasserstein GAN (WGAN) was developed to address the extreme class imbalance. By generating realistic synthetic crash instances, WGAN achieved higher sensitivity in crash detection while maintaining a low false alarm rate. Different from approaches using GAN, Zhang et al.~\cite{zhang2022accelerated} proposed an enhanced accelerated testing method based on Normalizing Flows. This method efficiently learned the distribution of rare crash events and preserves the correlations between scenario variables, significantly reducing the number of required tests by over two hundred times compared to traditional Monte Carlo simulations. However, the method’s effectiveness is highly dependent on the quality of the learned distributions and may require extensive computational resources for training the Normalizing Flows models.

\subsection{Road Segment-related Traffic Prediction}
Road segment-related traffic prediction focuses on forecasting traffic conditions (e.g., traffic flow, speed, travel time) for specific segments or sections of roads within a transportation network. Accurate prediction faces three main challenges. First, traffic conditions within the same road segment can vary over time, but there is often similarity between adjacent time periods. Additionally, the traffic state of one road segment can affect neighboring segments, resulting in complex spatial and temporal patterns in traffic conditions. Second, the limited number of road sensors and the possibility of sensor malfunctions can result in missing or inaccurate data, posing challenges for ensuring data quality and completeness. Third, external factors like accidents and weather further enhance the uncertainty of prediction. 
Generative AI technology has gained significant attention from scholars due to its ability to learn data distributions and model uncertainties. Numerous models based on generative AI technology are proposed for predicting traffic flow and travel time.

\textbf{Traffic flow prediction}. The task of traffic flow prediction is to learn a non-linear function $h(\cdot)$ from previous $T$ time steps of traffic conditions to forecast traffic conditions for the next $T^{'}$ steps based on road network graph $G$. 
Many scholars employ different generative AI techniques to tackle this task. We summarize the related works in Table~\ref{tab:segement_traffic_prediction}.

\begin{table}[t!]
\scalebox{0.7}{
\begin{tabular}{|p{2cm}|l|l|l|}
\hline
\textbf{Task}                                      & \textbf{Method}               & \textbf{Spatio-temporal modeling}                           & \textbf{Papers}                         \\ \hline
\centering \multirow{11}{*}{\textbf{\shortstack{Traffic flow \\prediction}}} & VAE                  & Temporal: LSTM                                     & \cite{jin2022pfvae}                   \\ \cline{2-4} 
                                          & \multirow{7}{*}{GAN} & Temporal: LSTM                                     & \cite{xu2020road}                    \\ \cline{3-4} 
                                          &                      & Spatial: GCN                                       & \cite{yu2020forecasting,zheng2022gcn} \\ \cline{3-4} 
                                          &                      & Spatial: GCN, Temporal: RNN                        & \cite{jin2022gan}                     \\ \cline{3-4} 
                                          &                      & Spatial: GCN, Temporal: GRU and self-attention     & \cite{khaled2022tfgan}                \\ \cline{3-4} 
                                          &                      & Spatial: GCN, Temporal: TCN                        & \cite{li2022mgc}                      \\ \cline{3-4} 
                                          &                      & Saptial: GraphSAGE; Temporal: LSTM                 & \cite{zhao2022graphsage}              \\ \cline{3-4} 
                                          &                      & Spatial: GCN; Temporal: attention network          & \cite{devadhas2023generative}         \\ \cline{2-4} 
                                          & Normalizing   flows  & Autoregressive network with convolution operations & \cite{zand2023flow}                   \\ \cline{2-4} 
                                          & \multirow{2}{*}{DPM} & Spatial: GCN, Temporal: TCN                        & \cite{wen2023diffstg}                 \\ \cline{3-4} 
                                          &                      & Temporal: RNN                                      & \cite{rasul2021autoregressive}        \\ \hline
\centering \multirow{2}{*}{\textbf{\shortstack{Travel time \\estimation}}} & GAN                  & Spatial: GCN                                       & \cite{song2021learn}                  \\ \cline{2-4} 
                                          & DPM                  & Masked Vision Transformer                          & \cite{lin2023origin}                  \\ \hline
\end{tabular}}
\caption{Related works about road segment-related traffic prediction.}\label{tab:segement_traffic_prediction}
\vspace{-12mm}
\end{table}

\emph{VAE}. Some researches have proposed VAE-based models. The authors in~\cite{boquet2020variational} designed a VAE-based prediction model with the ability to learn the data distribution, resulting in improved forecasting performance and versatility across multiple real-world datasets.
PFVAE~\cite{jin2022pfvae} presented a novel planar flow-based VAE for traffic flow prediction. By integrating LSTM as the autoencoder and employing planar flows within the VAE structure, the model effectively captured the intricate temporal dependencies and nonlinear patterns of time-series data. Empirical results demonstrated its robustness and accuracy in handling noisy time-series data.

\emph{GAN}. GAN-based models are commonly adopted in this task~\cite{lin2018pattern,zhang2019gcgan,saxena2019d,zhang2021satp,devadhas2023generative}. 
For example, ForGAN~\cite{koochali2019probabilistic} employed GAN to learn the underlying data distribution for accurate probabilistic forecasts. This approach not only enhanced prediction accuracy but also provided a robust framework for handling real-world fluctuations in time-series data. 
In particular, many researchers designed several methods to model spatiotemporal correlations in GAN-based prediction model~\cite{zang2019traffic,xu2020road,yu2020forecasting,zheng2022gcn,jin2022gan,khaled2022tfgan,li2022mgc,zhao2022graphsage}.
For example, PL-WGAN~\cite{jin2022gan} utilized GCN, RNN and attention mechanisms within a Wasserstein GAN framework to model spatiotemporal correlations in urban traffic networks, leading to highly accurate short-term traffic speed predictions. However, the combination of multiple neural network components increased computational demands.
Similarly, ASTGAN~\cite{liu2022attention} used the attention mechanism and an mask graph convolutional recurrent network to characterize temporal and spatial dependencies, improving data completeness and prediction accuracy. The GAN component ensured the reliability of forecasts, showing significant improvements in real-world traffic prediction tasks.


\emph{Normalizing flows}. Consider the capability of conditional normalizing flows in representing complex distributions characterized by high dimensionality and strong interdimensional relationships. MotionFlow~\cite{zand2023flow} utilized conditional normalizing flows to model complex spatio-temporal data distributions. This model effectively captured the intricate variability inherent in traffic patterns through its autoregressive conditioning on spatio-temporal inputs.

\emph{DPM}. Since DPM can effectively handle uncertainties within spatio-temporal graph neural network (STGNN), DiffSTG~\cite{wen2023diffstg} combined the spatiotemporal learning capability of STGNN with the uncertainty measurement of DPM, resulting in higher prediction accuracy.
The authors in~\cite{rasul2021autoregressive} developed an autoregressive denoising diffusion model for multivariate probabilistic time series forecasting, utilizing gradient estimation and Langevin sampling to accurately sample data distribution. Experimental results showed that it outperformed traditional and contemporary forecasting models in both accuracy and robustness.

\textbf{Travel time estimation}. Travel time estimation task aims to estimate the duration it takes to travel between two locations or along a specific route given the departure time. Several scholars employ generative AI techniques to address this task~\cite{song2021learn,lin2023origin}.  To be specific,
GCGTTE~\cite{song2021learn} integrated GCN and GAN to estimate travel time in a probabilistic form. This approach not only enhances prediction accuracy but also provides valuable insights into traffic variability, which is essential for proactive traffic management. 
Lin et al.~\cite{lin2023origin} proposed a two-stage diffusion-based origin-destination (OD) travel time estimation model. The model incorporated a conditioned pixelated trajectory denoiser to capture correlations between OD pairs and their historical trajectories, effectively managing variability and noise in trajectory data. It outperformed baseline methods in terms of accuracy, scalability, and explainability.

\subsection{Region-specific Traffic Prediction}
Region-specific traffic prediction focuses on forecasting traffic conditions such as traffic flow within specific geographic regions in a transportation network. The main challenges in achieving accurate predictions include complex spatio-temporal dependencies, limited data, and noise. Recent works~\cite{li2020gacnet,zhang2020off,zhang2019trafficgan,naji2021forecasting,zhang2023spatiotemporal,feng2021short} have aimed to employ generative AI techniques to tackle these challenges.
To be specific, 
Kang et al.~\cite{kang2020generative} developed a GAN-based model that integrates gated and dilated convolutions to effectively capture both local and distant traffic patterns within urban regions. This dual convolution approach allows for more precise real-time traffic flow predictions compared to traditional grid-based models.
Li et al.~\cite{li2022spatial} proposed a seq2seq spatial-temporal semantic GAN model for multi-step urban flow prediction. By treating successive urban flow data as video frames, the model employed a spatial-temporal semantic encoder to simultaneously capture semantic factors and spatial-temporal dependencies. The integration of adversarial loss with prediction error effectively mitigates the issue of blurry predictions, resulting in more accurate and reliable forecasts. 
The authors in~\cite{naji2021forecasting} incorporated recurrent and conventional network models into the GAN framework to predict the taxi demand in specific areas. The model demonstrated superior performance on real-world data, and provided valuable insights into the dynamics of taxi supply and demand.


\subsection{Discussion}

Generative AI has made great progress in traffic prediction, addressing challenges in data quality, behavioral modeling, and spatiotemporal dynamics. Despite their high accuracy, these models often lack interpretability, which is crucial for building trust in ITS. 
Approaches like DVAE~\cite{neumeier2021variational}, which incorporate expert knowledge, improve interpretability without sacrificing performance. This balance is crucial for gaining trust and facilitating the adoption of generative AI technology in ITS.
Real-time applications also present computational challenges. While GPT-based frameworks~\cite{su2022crossmodal,feng2023trtr,vishnu2023improving} offer high prediction accuracy, they struggle with latency and resource demands, highlighting the need for optimized models that balance performance and efficiency.
In summary, achieving a balance between accuracy, interpretability, and efficiency remains critical in traffic prediction. Future research should focus on developing generative AI models that not only achieve high accuracy but also maintain transparency and efficiency in real-time traffic environments. 
\section{Generative AI for Traffic Simulation}\label{sec:paradigms}
Traffic simulation involves the mathematical modeling of transportation systems, which can be used to generate the movement and behavior of vehicles and pedestrians. This technology can be applied in various fields, including urban planning, evaluation of policy changes, and testing and validating autonomous vehicle algorithms. However, achieving a more realistic traffic simulation poses several challenges. First, real-world traffic scene data is typically collected through road cameras or sensors. However, obtaining such data is not only costly but also challenging to capture in rare or hazardous situations. Second, traffic dynamics are inherently complex due to the numerous variables involved, ranging from individual driver behaviors, decisions, and reactions to the unpredictable nature of external factors such as weather or sudden events. Capturing the intricacies of these dynamics is a considerable challenge. Additionally, the interconnected relationships between vehicles, infrastructure, and pedestrians further complicate the situation. Existing works using traditional deep learning methods~\cite{halim2016profiling,xiao2022perception} often require extensive labeled data, and may not generalize well to unseen or rare situations. In addition, these methods may struggle to capture the vast array of dynamic interactions in traffic scenarios~\cite{wei2014behavioral,li2021safe}. 

\begin{figure}[t!]
	\centering
	\includegraphics[width=4in]{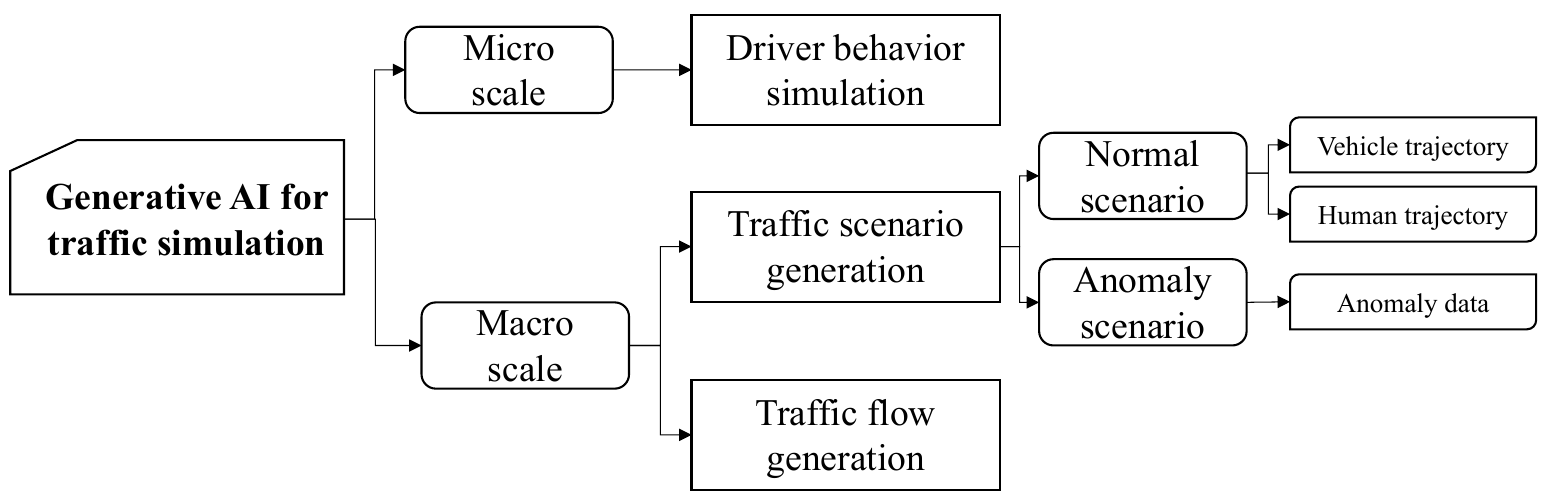}
 \vspace{-3mm}
	\caption{The relation between different topics in generative AI for traffic simulation research.}
	\label{fig:traffic_simulation}
	\vspace{-6mm}
\end{figure}

Generative AI techniques offer potential solutions for traffic simulation challenges. They can create realistic traffic scenarios, including rare or extreme situations that might be missing from real-world data. Their adaptability allows them to adjust to ever-changing traffic dynamics, thereby capturing intricate interactions more holistically. Furthermore, by learning from a small number of real traffic data, they can refine their simulations continuously, ensuring they mirror the complexities of real-world traffic more closely. Next, we will introduce the related works using generative AI techniques from three aspects: driver behavior simulation, traffic scenario generation and traffic flow generation. We draw their relation in Figure~\ref{fig:traffic_simulation}.

\subsection{Driver Behavior Simulation}

In autonomous driving scenarios, the focus is on how drivers make decisions in response to traffic conditions~\cite{ghosh2016sad,yun2019data,chen2022combining,ye2022efficient,singh2023bi}. 
Yun et al.~\cite{yun2019data} utilized GAN to develop a data-driven, human-like driver model capable of simulating intricate interactions in traffic scenarios. By leveraging GANs, their model effectively captured the intricate driving skills exhibited by humans in diverse and complex situations, outperforming traditional rule-based models.
Chen et al.~\cite{chen2022combining} introduced a two-stage hybrid driver model that integrates data-driven neural networks with model-based controllers. By employing reward-augmented GAIL, their model effectively simulated human-like driving behaviors while ensuring safety and expressiveness in multi-agent highway scenarios. 
In a novel application of LLM, Jin et al.~\cite{jin2023surrealdriver} developed a generative driver agent simulation framework that leverages human driver experience through self-reported driving thoughts. By integrating LLMs with human-derived chain-of-thought data, their framework significantly reduces collision rates and enhances the human-like qualities of simulated drivers. 
Addressing the computational challenges associated with large-scale simulations, Ye et al.~\cite{ye2022efficient} proposed an efficient calibration method for agent-based traffic simulations using VAEs. By compressing agent state vectors into lower-dimensional latent spaces, their approach dramatically reduced the computational burden of calculating state transfer probabilities.

Lane changing and car following are two common driving behaviors, and numerous studies have explored methods for simulating these behaviors~\cite{lin2021car,ma2023physics,li2023operational,dong2023transfusor}. 
Ma et al.~\cite{ma2023physics} introduced a physics-informed conditional GAN that integrates the advantages of both physics-based models and deep learning techniques. This hybrid approach eliminated the need for explicit weighting parameters, simplifying the model while effectively capturing multi-step car-following behaviors in complex traffic environments. 
Lin et al~~\cite{lin2021car} introduced a constrained-GAIL framework that employs reward augmentation to guide driving agents away from  undesirable behaviors. By incorporating manually designed reward functions alongside those learned from data, their model achieved higher prediction accuracy in vehicle speed and location while significantly reducing collisions and erratic driving behaviors. 
Addressing the need for highly realistic simulation environments, the authors in~\cite{dong2023transfusor} leveraged the transformer and diffusion models to generate human-like lane-changing trajectories. Extensive experiments validated its ability to produce realistic and controllable trajectories.

Unlike autonomous driving scenarios, in transportation services, a variety of factors such as estimated travel demand tend to dominate drivers' behaviors. For example, Zhang et al.~\cite{zhang2020cgail} developed conditional GAIL model capable of learning the driver’s decision-making tendencies and strategies through knowledge transfer across different drivers and locations. By integrating collective inverse RL, cGAIL enhances the generalization of driver behaviors, thereby improving traffic management and service quality.

\subsection{Traffic Scenario Generation}

\begin{table}[t!]
\scalebox{0.8}{
\begin{tabular}{|p{3cm}|l|l|l|}
\hline
\multicolumn{1}{|l|}{\textbf{Task}}                     & \textbf{Method}   & \textbf{Controllability} & \textbf{Papers}                                                                                         \\ \hline
\centering \multirow{6}{*}{\textbf{\shortstack{Vehicle trajectory\\ generation}}} & VAE      & Unconditional   & \cite{chen2021trajvae,zhang2022factorized}                                                    \\ \cline{2-4} 
                                               & GAN      & Unconditional   & \cite{kumar2023generative,liao2022itgan,ozturk2020development,liao2022itgan,choi2021trajgail} \\ \cline{2-4} 
                                               & DPM      & Unconditional   & \cite{sun2023drivescenegen,zhu2023diffusion,xu2023diffscene,xu2023generative}                 \\ \cline{2-4} 
                                               & VAE, GAN & Unconditional   & \cite{krajewski2018data,krajewski2019beziervae}                                               \\ \cline{2-4} 
                                               & GAN      & Conditional     & \cite{arnelid2019recurrent}                                                                   \\ \cline{2-4} 
                                               & DPM      & Conditional     & \cite{zhong2023language,zhong2023guided}                                                      \\ \hline
\centering \multirow{2}{*}{\textbf{\shortstack{Human trajectory \\ generation}}}   & GAN      & Unconditional   & \cite{xiong2023trajsgan}                                                                      \\ \cline{2-4} 
                                               & DPM      & Conditional     & \cite{rempe2023trace}                                                                         \\ \hline
\centering \multirow{3}{*}{\textbf{\shortstack{Anomaly data \\generation}}}       & VAE      & Unconditional   & \cite{cai2020real,islam2021crash}                                                             \\ \cline{2-4} 
                                               & GAN      & Unconditional   & \cite{chen2021trafficacc,chen2022efficient,huo2021text}                                       \\ \cline{2-4} 
                                               & GAN      & Conditional     & \cite{zarei2021crash}                                                                         \\ \hline
\end{tabular}}
\caption{Related works about traffic scenario generation.}\label{tab:traffic_scenario_gen}
\vspace{-12mm}
\end{table}

Traffic scenario generation refers to produce specific traffic situations or anomalous events for the purpose of testing and validating autonomous driving systems. As autonomous vehicles need to operate safely in a vast array of driving conditions, it is crucial for developers to test these systems against a wide range of potential real-world scenarios, especially those that are rare or potentially hazardous. In this literature review, we will focus on three key areas: generating vehicle trajectories, generating human trajectories, and generating anomaly data, as summarized in Table~\ref{tab:traffic_scenario_gen}.

\textbf{Vehicle trajectory generation}.
Many researchers use generative AI technology to generate more vehicle trajectory data~\cite{krajewski2018data,krajewski2019beziervae,choi2021trajgail,liao2022itgan,zhang2022factorized,chen2021trajvae,ozturk2020development,kumar2023generative,sun2023drivescenegen,xu2023diffscene}.
In~\cite{krajewski2018data}, the authors introduced trajectory GAN and trajectory VAE to generate realistic synthetic lane-change maneuver trajectories without expert knowledge. This approach not only facilitated the creation of diverse training datasets for autonomous driving systems but also reduced the dependency on extensive expert annotations.
PS-TrajGAIL~\cite{wang2024urban} modeled urban vehicle trajectory generation as a partially observable Markov decision process within a GAIL-based structure. This approach leveraged IL to capture the complex decision-making processes of human drivers, thereby generating synthetic trajectories that exhibited high fidelity to real-world behaviors. Experimental evaluations on both synthetic and real-world datasets demonstrated that PS-TrajGAIL significantly outperforms existing baselines in terms of trajectory realism and predictive accuracy.
Diff-Traj~\cite{zhu2023diffusion} was designed for generating GPS trajectories by learning spatio-temporal features from historical trajectories. By employing DPMs, this model captured the stochastic and dynamic nature of human driving behaviors, producing high-quality synthetic trajectories that retain the statistical properties of real-world data. 
WcDT~\cite{yang2024wcdt} combined DPMs with transformers to generate diverse and realistic autonomous driving trajectories. By encoding historical data and environmental features through transformer-based encoders, WcDT effectively captured multifaceted interactions and temporal dependencies present in driving scenarios. The integration of diffusion models enhanced the framework’s ability to produce high-fidelity trajectories that reflect the variability and unpredictability of human driving behavior. 

Differently, some searchers focused on the controlled vehicle trajectory generation~\cite{arnelid2019recurrent,zhong2023language,zhong2023guided}. 
For example, the authors in~\cite{arnelid2019recurrent} proposed RC-GAN, a recurrent conditional GAN model that incorporated LSTM in both the generator and discriminator, enabling the generation of sensor errors with long-term temporal correlations. Empirical evaluation on real road data demonstrated that RC-GAN produced significantly more realistic sensor error simulations.
Zhong et al.~\cite{zhong2023language} introduced a scene-level conditional diffusion model that incorporated a spatio-temporal Transformer architecture to produce realistic and controllable traffic data. A key innovation was incorporating language instructions through a LLM, which translated user queries into loss functions that guide the diffusion model. Evaluations demonstrated that this approach not only produced realistic traffic patterns but also effectively adhered to user constraints, outperforming existing methods in both realism and controllability.
Xu et al.~\cite{xu2023generative} designed a generative AI-based autonomous driving architecture that utilizes large text-to-image models based on DPM to generate conditioned traffic and driving data in simulations. The approach improved simulation fidelity, providing a robust tool for testing and validating autonomous driving systems.

\textbf{Human trajectory generation}. 
Regarding human trajectory generation, Xiong et al.~\cite{xiong2023trajsgan} proposed a semantic-guiding GAN model that integrates semantic knowledge to generate realistic human trajectories. By employing an attention-based generator and a CNN-based discriminator, this approach effectively captured both the sequential and spatial dependencies of human movements.  Experiments validated its utility in simulating epidemic spread scenarios, achieving high correlation with real data.
STAGE~\cite{cao2024stage} employed a multi-task GAN with spatiotemporal knowledge, integrating a multi-stage generator and spatial consistency loss to produce privacy-preserving synthetic trajectory. Experimental results indicated that STAGE outperformed baseline models in data distribution alignment, making it a robust tool for applications requiring high data fidelity and privacy protection. 
G-GAIL~\cite{li2024game} integrated game theory with GAIL to model pedestrian-vehicle interactions as a Markov decision process, enabling realistic simulation of pedestrian behaviors and enhancing autonomous driving scenario testing.  The model demonstrated superior performance in replicating pedestrian behaviors under various conditions, enhancing the reliability of autonomous driving scenario testing. 
Rempe et al.~\cite{rempe2023trace} introduced a controlled pedestrian trajectory generation approach using guided diffusion modeling and a physics-based humanoid controller, generating realistic movements based on user-defined goals and environmental constraints, thus improving the realism and flexibility of simulations. 

\textbf{Anomaly data generation}.
Anomaly data such as traffic accident and crash data is essential in autonomous driving for safety assessment and scenario testing.
However, this type of traffic data is hard to collect. For example, as illustrated in~\cite{cai2020real}, the ratio of crash data to non-crash data is 1 to 11,000, indicating a significant imbalance. Several works use generative AI techniques to address this problem~\cite{islam2021crash,chen2021trafficacc,chen2022efficient,huo2021text,zarei2021crash}.
For instance, Islam et al.~\cite{islam2021crash} employed VAE to encode imbalanced crash and non-crash events into a latent space for generating realistic crash data. This approach not only achieved statistical similarity with real crash data but also significantly improved prediction model performance.
In~\cite{huo2021text}, a text-to-traffic GAN model was proposed, which integrates traffic data with semantic information collected from social media. By incorporating a global-local loss to bridge the modality gap between textual and traffic data, their model achieved superior authenticity in generated traffic scenarios.
Zarei et al.~\cite{zarei2021crash} developed a conditional GAN-based model that utilizes crash counts as conditions to generate realistic crash data. Their method demonstrated superior performance in hotspot identification, prediction accuracy, and dispersion parameter estimation, especially in low-dispersion scenarios.

\subsection{Traffic Flow Generation}
Traffic flow generation refers to generate vehicular traffic volumes or crowd flow within a specific area or between regions in a transportation network. Numerous works adopt generative AI techniques to address this task~\cite{chen2021traffic,li2022attentive,zhou2023towards,rong2023complexity,rong2023origin}.
Specifically, 
Li et al.~\cite{li2022attentive} presented an attentive dual-head spatial-temporal GAN model for generating crowd flow data.
This model incorporated an attentive mechanism for tracking temporal changes and a self-attention network to capture intricate spatiotemporal crowd flow patterns, and it adopted a dual-head discriminator that implements a training strategy with two distinct objectives to mitigate the adverse effects of rapid overfitting.
Zhou et al.~\cite{zhou2023towards} developed a knowledge-aware spatio-temporal diffusion model that utilizes an urban knowledge graph to generate dynamic urban flow data. By integrating region features and environmental factors, this model accurately generated urban flows for regions without historical data.
In~\cite{rong2023complexity}, the authors employed a graph denoising diffusion technique to generate urban OD flow, effectively capturing the interconnections of nodes and edges within large urban networks. By decomposing the generative process into network topology and edge weight generation, their model achieved superior realism in OD matrices across multiple cities.

\subsection{Discussion}
The literature review highlights several key lessons regarding the application of generative AI techniques in traffic simulation.
Firstly, generative models in~\cite{zarei2021crash,islam2021crash} effectively address data imbalance by synthesizing realistic crash and rare event data. Secondly, integrating multi-modal and semantic information, as seen in $\text{T}^2$GAN~\cite{huo2021text}, enriches traffic scenario generation by combining traffic data with social media insights, resulting in more authentic and contextually relevant simulations. Moreover, privacy-preserving techniques like STAGE~\cite{cao2024stage} ensure data fidelity while protecting individual privacy, making synthetic data suitable for sensitive applications. However, these advanced models often face challenges related to computational complexity and scalability, highlighting the need for further optimization to enable real-time and large-scale implementations.

\section{Generative AI for Traffic Decision-making}

Traffic decision-making involves a series of decision-making processes made by vehicles or other road users based on dynamic traffic environment and specific task requirements. Among these processes, route planning, traffic signal control (TSC) and vehicle control are three fundamental aspects. 
Route planning generally refers to generating an optimal path from a designated starting point to a destination, considering factors such as departure time, road conditions, and traffic patterns. 
TSC plays a crucial role in coordinating the movement of vehicles and pedestrians at intersections, thereby reducing traffic congestion.
Vehicle control involves managing the vehicle’s operation, including actions such as acceleration, steering, and braking, to navigate safely along driving route.

\begin{figure}[t!]
	\centering
	\includegraphics[width=3.5in]{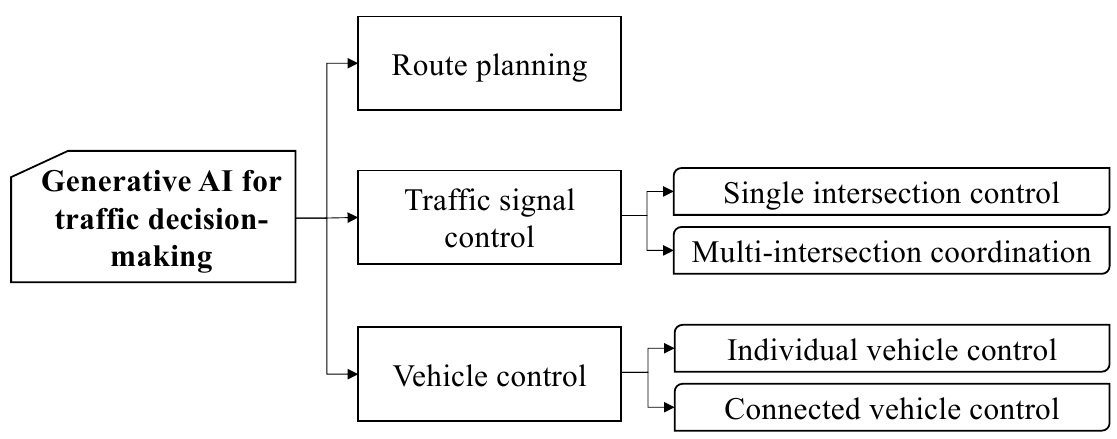}
 \vspace{-3mm}
	\caption{The relation between different topics in generative AI for traffic decision-making research.}
	\label{fig:traffic_decision}
	\vspace{-6mm}
\end{figure}

Ensuring that route planning, TSC, and vehicle control collaboratively make safe and accurate decisions in complex environments is a formidable challenge.
One significant challenge in traffic decision-making lies in the unpredictable dynamics of traffic environments. These environments are constantly changing due to various factors, including road conditions, unexpected moves by human drivers, and sudden decisions by pedestrians and cyclists. For example, a driver might suddenly brake or change lanes without warning, or pedestrians might cross the road unexpectedly. These uncertainties compel autonomous systems to be perpetually alert and adaptive, which becomes computationally intensive and requires robust prediction algorithms.
Another critical challenge involves rare but critical events that can pose significant risks to safety, such as a child running onto the road or a mechanical failure causing loss of control. Collecting real-world data on these rare events is difficult but crucial for creating a comprehensive training dataset. Without exposure to these scenarios, autonomous systems may not know how to handle them properly, potentially leading to safety risks. 
Additionally, the limitations of sensors (e.g, LiDAR, RADAR, and cameras) present another obstacle. These sensors often have restrictions regarding their operational range and resolution, as well as their susceptibility to adverse weather conditions, including fog, rain, and snow. Such limitations can introduce significant uncertainties, resulting in suboptimal decision-making and compromising both safety and efficiency within traffic operations.

Traditional decision-making methods, such as rule-based methods~\cite{mcnaughton2011motion,de2017decision} and game theory-based methods~\cite{wang2015game,hang2020human,zhao2021yield}, often have limited effectiveness in dealing with complex traffic situations, while reinforcement learning (RL), as a flexible, efficient, and powerful method, has gained ground in traffic decision-making. However, the training complexity associated with handling multiple tasks limits its widespread applicability. 
Generative AI technology offers promising solutions to these challenges by leveraging its strengths in data generation, reasoning, and sequence modeling.
For instance, its sequence modeling capabilities help in understanding and predicting complex driving patterns over time. 
Furthermore, generative AI’s strong analytical and reasoning abilities allow for more effective interpretation of external traffic conditions and instructions, thereby providing important guidance for accurate decision-making. 
To mitigate the impact of rare events, generative AI can generate realistic training scenarios mimicking the unpredictability of real-world driving, reducing the need for extensive road testing. By producing rare and critical traffic events, generative AI ensures autonomous systems are robust against challenging situations.
Furthermore, generative AI can enhance sensor data, enabling vehicles to handle challenging environmental conditions with greater adaptability and accuracy.

In the following subsections, we conduct a comprehensive literature review and explore how generative AI can enhance decision-making processes within traffic systems, specifically focusing on route planning, traffic signal control and vehicle control. Their relationship is illustrated in Figure~\ref{fig:traffic_decision}.


\subsection{Route Planning} 

Generative approaches have become integral to route planning tasks within intelligent transportation systems, leveraging their robust data generation and powerful learning capabilities to incorporate global geographical knowledge. 

One category of approaches utilizes GANs to produce realistic and efficient routes~\cite{fu2021progrpgan,choi2022pathgan,wang2022data}. 
For example, ProgRPGAN~\cite{fu2021progrpgan} employed a progressive route planning GAN to create realistic paths across multi-level map resolutions. 
Similarly, Choi et al.~\cite{choi2022pathgan} developed a GAN-based framework that integrates feature extraction and path generation networks to produce multiple plausible routes from egocentric images without relying on high-definition maps.

Another category of approaches integrates RL with generative models to achieve expert-level path planning. 
MixGAIL~\cite{yang2024mixed} combined transition-aware adversarial imitation learning and GAIL techniques within a mixed policy gradient actor-critic framework.
Building on this, a VAE-GAN framework for path optimization~\cite{wang2022varl} combined VAEs with GANs to address the vehicle routing problem on graphs. This approach utilized GNNs to model the complex relationships within road networks and employed variational reasoning to identify optimal nodes by dividing the graph into subgraphs centered around a root node. By integrating RL, the framework achieved end-to-end optimization, enhancing both learning efficiency and generalization capabilities.

In addition, the integration of LLMs represents a great advancement in providing explainable and adaptive path planning. 
DynamicRouteGPT~\cite{zhou2024dynamicroutegpt} utilized causal inference and pretrained language models to offer explainable and adaptive real-time path planning in complex traffic scenarios. 
Addressing the limitations of traditional algorithms in large-scale environments, LLM-A*~\cite{meng2024llm} proposed a novel path planning method that synergistically combines the precise pathfinding capabilities of A* with the global reasoning abilities of LLMs. This hybrid approach aims to enhance pathfinding efficiency in terms of time and space complexity while maintaining path validity.

\subsection{Traffic Signal Control}

Generative methods greatly improve the effectiveness of TSC by improving data integrity, scalability, and decision-making efficiency.
The existing literature on generative methods for TSC can be divided into two categories: single-intersection and multi-intersection control approaches.

\textbf{Single-intersection control.}
Generative approaches improve single-intersection traffic signals by recovering missing data. For instance, DiffLight~\cite{chen2024difflight} addressed missing data in TSC by combining data imputation with decision-making, using a diffusion model to ensure robust control even with incomplete data.
Other research focuses focus on the integration of LLMs with RL~\cite{wang2024llm,pang2024illm}. The authors in~\cite{wang2024llm} introduced a hybrid framework that integrates LLMs with perception and decision-making tools, enabling the system to adapt to diverse traffic conditions without additional training. 
Pang et al.~\cite{pang2024illm} further advanced this integration by employing LLMs to evaluate and adjust RL decisions under degraded communication and rare events. These integrative methods demonstrate substantial advantages in adaptability and resilience, effectively addressing the limitations inherent in traditional RL-based systems when confronted with dynamic and unpredictable traffic environments.

\textbf{Multi-intersection coordination.} 
This category focuses on data recovery and the coordination of traffic signals across multiple intersections. A decentralized adaptive TSC framework~\cite{wang2021gan} employed a GAN-based algorithm for efficient traffic data recovery. This approach significantly reduced vehicle travel time and maintained high throughput by enabling collaboration with neighboring intersections without heavy data exchange. 
Similarly, Xu et al.~\cite{xu2024graph} leveraged a Wasserstein GAN to estimate agent state spaces for multi-intersection settings, which effectively handles missing data and enhances system robustness. When integrated with a graph-based deep RL model, this approach outperformed baseline models in handling complex interactions and incomplete data in multi-intersection environments.

\subsection{Vehicle Control}
Vehicle control primarily involves exploring decision-making approaches in autonomous driving scenarios, including individual vehicle control and connected vehicle control strategies.

\textbf{Individual vehicle.}
 This subsection discusses methods to enhance decision-making for individual vehicles using LLM-assisted approaches and other generative techniques.
 
\emph{LLM-assisted approaches.} The integration of LLMs into autonomous vehicle systems has emerged as a transformative approach to enhancing decision-making capabilities in data analysis, reasoning, and sequential decision modeling. By leveraging LLMs, these systems can effectively interpret and process multi-modal external data, perform complex reasoning akin to human commonsense, and generate coherent decision sequences that enhance navigation and safety.
For instance, DriveGPT4~\cite{xu2024drivegpt4} and TrafficGPT~\cite{zhang2024trafficgpt} illustrated how LLMs can process multi-frame video inputs and traffic data to produce vehicle controls and decompose complex tasks, respectively. This capability significantly enhances the system’s ability to manage intricate driving scenarios.
Additionally, frameworks proposed in works~\cite{cui2024receive} and~\cite{cui2024drive} incorporated environmental information and driver commands, enabling human-like reasoning and real-time optimization of driving maneuvers. These frameworks also personalized user experiences and foster trust through explainable decisions.
Furthermore, cognitive reasoning pathways and translation algorithms, as employed in the literature~\cite{sha2023languagempc}, ensured seamless integration of high-level LLM decisions with low-level vehicle controls, thereby enhancing adaptability and reliability. 
Moreover, models such as the multi-task decision-making GPT~\cite{liu2023mtd} and the sequence modeling approach within a constrained Markov decision process~\cite{liu2024decision} explored the capability of LLMs to handle multiple driving tasks simultaneously and improve exploration and safety through advanced techniques like entropy regularization.
The use of graph-of-thought structures in explainable decision-making~\cite{chi2024multi} highlighted the importance of transparency and trust by generating natural language rationales for driving choices. 
In summary, this integration not only addresses critical challenges such as safety assurance, generalization to rare events, and interpretability but also significantly improves the safety, efficiency, and reliability of autonomous driving, paving the way for more intelligent and trustworthy transportation solutions.

\emph{Other generative techniques.} Besides LLM-aided approaches, other generative techniques provide great advancements in autonomous vehicle decision-making by enhancing robustness and safety in complex driving scenarios~\cite{couto2024hierarchical,liu2024ddm,yang2024diffusion}.
For example, the authors in~\cite{couto2024hierarchical} propose a multi-level GAIL framework that uses a GAN to generate Bird’s-Eye View representations and a GAIL module to learn vehicle control from these representations, effectively decoupling representation learning from the driving task to achieve robust autonomous navigation. 
DDM-Lag~\cite{liu2024ddm} used a diffusion-based generative process to model autonomous vehicle decision-making, incorporated Lagrangian-based safety enhancements, and integrated hybrid policy updates with safety-constrained optimization to enhance decision performance in complex driving scenarios. This approach addresses the critical need for safety assurance in AVs by mitigating risks associated with decision-making under uncertainty, ensuring reliable performance in intricate driving situations. 

\textbf{Connected vehicles.}
Recent advancements in connected vehicle control demonstrate the significant benefits of integrating LLMs into autonomous driving frameworks. CoMAL~\cite{yao2024comal} leveraged LLMs to facilitate collaboration among autonomous vehicles, enhancing traffic flow and reducing congestion. A key component of CoMAL is the reason engine, which processes scenario descriptions and system prompts as inputs. Using a hierarchical chain-of-thought prompt, the LLM generated driving strategies based on the intelligent driver model, allowing each vehicle to respond effectively to dynamic situations. This approach not only outperformed traditional reinforcement learning methods on the Flow benchmark but also highlighted the strong cooperative capabilities of LLM-driven agents.
Similarly, CoDrivingLLM~\cite{fang2024towards} employed LLMs within a reasoning module for decision-making, incorporating state perception, intent sharing, negotiation, and chain-of-thought processing to produce final actions. Experimental results showed that CoDrivingLLM surpasses rule-based, optimization-based, and machine learning methods, achieving higher success rates, effective conflict resolution, and balanced efficiency in complex multi-vehicle traffic scenarios.

\subsection{Discussion}
While existing literature has made significant progress in using generative AI to enhance the effectiveness of traffic decision-making, there are still areas for improvement in capturing the unpredictable dynamics of traffic environments, addressing rare but critical events, and overcoming the limitations of sensors. 
For instance, current LLM-aided methods such as DriveGPT4~\cite{xu2024drivegpt4} and TrafficGPT~\cite{zhang2024trafficgpt}, although capable of processing multi-frame video and traffic data, still require enhancements in decision robustness and real-time responsiveness in unknown scenarios. Additionally, while methods like DiffusionES~\cite{yang2024diffusion} attempt to augment the flexibility and diversity of autonomous vehicle decision-making strategies, the diversity and authenticity of generating extremely rare events remain to be validated. Furthermore, the literature provides limited discussion on enhancing perception and decision-making capabilities under adverse environmental conditions. Addressing it will be a crucial area for future research.

\section{Open Challenges and Future Directions}\label{sec:challenges}

While generative AI methods have demonstrated much success in the realm of intelligent transportation systems, several challenges must be addressed. In this section, we will discuss the primary open challenges, followed by an introduction to some potential research directions.

\subsection{Open Challenges}

In this subsection, we will discuss several open challenges for generative AI methods in intelligent transportation systems.

\textbf{Multi-modal traffic data}.
Multi-modal traffic data combines various sources, such as roadside camera videos, traffic signals, weather conditions, incident reports, and vehicle GPS trajectories, providing a comprehensive view of traffic conditions. However, using generative AI to solve traffic-related problems with multi-modal data presents challenges. 
Firstly, the correlations between multi-modal data are more complex compared to single-modal data, as the characteristics and distributions of images, text, and other formats differ significantly. Secondly, accurate alignment between modalities, such as matching traffic camera images with sensor or weather data, is critical. Additionally, training models with multi-modal data can be unstable, especially when maximizing cross-modal information. Therefore, how to effectively use generative AI technology to tackle transportation challenges using multi-modal data remains challenging in the field of transportation.

\textbf{Complex spatio-temporal correlations}.
Traffic involves complex relationships across time, space, and spatiotemporal interactions. In the temporal dimension, traffic data is influenced by factors like weekdays, weekends, and holidays, and current traffic flow can be impacted by patterns from the preceding hours. 
In the spatial dimension, traffic conditions at different locations are interrelated, requiring analysis of correlations between traffic flow and congestion across locations. Additionally, traffic flow at one location may be affected by nearby traffic in the recent past.
Previous research has employed RNNs, CNNs, and GNNs to model spatiotemporal relationships in traffic, showing promising results. However, most approaches focus on local or micro-scale spatiotemporal modeling and fail to capture global or macro-scale spatiotemporal patterns. Furthermore, spatiotemporal modeling has yet to fully utilize the powerful inference capabilities of generative AI. The main challenge lies in effectively modeling the intricate spatiotemporal correlations within traffic data to address traffic-related issues.

\textbf{Sparse or missing traffic data}.
Traffic data sparsity or gaps can arise due to equipment failures, maintenance, network issues, or the costs and privacy concerns of data collection. Such data gaps can lead to inaccurate traffic predictions and incomplete insights for traffic management. Generative AI, with its powerful data generation capabilities, has been used to address this issue through data imputation, traffic estimation, and scene generation. However, there are still shortcomings. 
First, due to the spatiotemporal complexity of traffic data, generative AI models may produce data that appears realistic but lacks spatial or temporal consistency. Second, when generating data, the generative model may tend to generate common patterns in training data while neglecting less common cases. This may lead to a lack of diversity in the generated data and inability to cover all the changes in traffic data. Third, different levels of data sparsity can significantly affect generation quality, requiring models to adapt flexibly.
Therefore, overcoming the challenges of data sparsity and absence in addressing traffic issues with generative AI technology remains an important problem that needs further research.

\textbf{Adversarial attacks}. 
In intelligent transportation, generative AI technology offers cutting-edge solutions for traffic management and optimization. However, its widespread adoption also introduces new security risks, particularly from adversarial attacks~\cite{foo2023aigc}. One major concern is the backdoor attack, where malicious data injected during training can cause models to produce harmful outputs under specific conditions.
In complex intelligent transportation networks, a single misjudgment malicious data injected during training can cause models to produce harmful outputs under specific conditions. For instance, an intelligent transportation monitoring system compromised by a backdoor attack may erroneously classify a safe vehicle as a potential threat, or an autonomous vehicle may make incorrect driving decisions as a result. 
Therefore, the critical challenge lies in accurately identifying potential risk points and devising effective methods to ensure the efficient and secure operation of transportation systems.

\textbf{Model interpretability}.
 Model interpretability is crucial in ITS. For example, in autonomous driving scenarios, predicting the behavior of other road users, such as pedestrians, cyclists, and vehicles, is crucial for safe driving. This requires the model to explain why a particular behavior is considered the most likely, such as why a pedestrian may cross the road. 
 While generative AI models excel in processing traffic data and optimizing transportation systems, they are often seen as black boxes, which makes it difficult to understand their decision-making. Designing interpretable generative AI models is challenging due to the complexity of traffic systems, involving numerous variables and uncertainties like road conditions, traffic flow, and pedestrian behavior. In autonomous driving, autonomous driving decisions involve multiple levels, including perception, planning, and control, and designing an interpretable model requires being able to explain the decision-making processes at these different levels, which are often very complex.
 
\textbf{Real-time requirements}.
In ITS, timely and effective response measures are crucial, especially in rapidly changing situations or unforeseen events. Traffic models must have real-time perception capabilities to detect vehicles, pedestrians, and obstacles while making immdiate decisions to ensure safety. Additionally, they must predict changes in traffic flow to take proactive measures against congestion and accidents.
However, achieving this high level of responsiveness poses some challenges in generative AI-based solutions. First, real-time data may be affected by noise and incompleteness, so models must have the ability to handle imperfect data to accurately perceive the traffic environment. Second, certain generative AI methods may be very complex and require more time for reasoning. In real-time applications, it is necessary to balance the complexity and real-time performance of the model. Third, real-time applications should be equipped to handle exceptional circumstances like sensor failures or sudden events. Generative AI methods need to be capable of identifying such anomalies and executing appropriate actions to ensure both safety and reliability.

\textbf{High-perfomance computing}.
In large metropolis, high population density results in diverse activities, complex travel behaviors, and numerous transportation routes, all of which present significant challenges for evaluating and optimizing traffic management in ITS~\cite{aboudina2018harnessing}. Further, leveraging generative AI for large-scale traffic simulations and optimizations involves running multiple simulations simultaneously, performing intensive analyses to find optimal solutions, and processing large volumes of data. High-Performance Computing (HPC) offers opportunities to provide the required computational power to meet these intensive demands. However, applying HPC to support generative AI in ITS presents several challenges. As the massive amounts of data generated from numerous sources in ITS applications, highly efficient data handling and parallel processing are crucial for enabling timely decision-making. Another key challenge is ensuring scalability and efficiency in real-time applications. ITS environments are highly dynamic, with traffic conditions changing rapidly due to factors such as accidents, weather changes, or special events. The HPC framework must be able to scale computational resources to handle sudden spikes in data volume or computational demand, ensuring the system remains responsive under varying conditions.

\subsection{Potential Research Directions}
Large-scale language models designed using generative AI technology have achieved significant success in fields such as NLP and image processing, which also provide crucial opportunities and research directions for addressing traffic-related challenges.  
In this subsection, we will introduce several important research directions from three distinct perspectives concerning large-scale traffic models. 

\emph{Integrating multi-modal data for large-scale traffic models}. Traffic data usually come from multiple different sources, including images and video data from traffic cameras, as well as GPS trajectory data from vehicles, covering a variety of different data types. These data offer multiple perspectives on the spatiotemporal dynamics of traffic conditions, facilitating a deeper comprehension of traffic operation's underlying mechanisms and patterns. Therefore, these data serve as an important foundation for training large-scale models in the field of transportation. Efficiently using these data for model training is crucial and also an important research direction. In addition to aligning the data in the spatiotemporal dimension, it is necessary to unify their representation, mapping different data modalities to the same data space. This requires designing appropriate mapping methods to ensure that the mapped data can effectively capture spatial and temporal dependencies across different data modalities.

\emph{Designing and training well-performed large-scale traffic models}.
Traditional transportation models are typically developed separately for specific tasks, requiring significant time and resources to adapt to new requirements. These models also tend to be overly specialized, making cross-task transfer learning difficult and limiting scalability.
In contrast, large-scale language models based on generative AI technology have higher generalizability and scalability, with the potential for transfer learning across various traffic tasks. For instance, in autonomous driving, which encompasses continuous tasks such as perception, prediction, and decision-making, a large-scale model could handle all tasks simultaneously by modeling shared features like spatiotemporal characteristics while considering each task’s unique aspects.
Recently, GPT-ST~\cite{li2023gpt} gives a preliminary attempt in designing generative spatio-temporal pre-training framework.
Another promising direction is developing large-scale traffic models driven by both data and knowledge, combining diverse traffic data with reliable physical knowledge to effectively model traffic patterns.s

\emph{Traffic planning and decision-making based on large-scale language models}.
Traffic planning and decision-making are important research problems in the transportation domain. In real-world scenarios, traffic conditions constantly change due to congestion, accidents, or adverse weather, impacting the effectiveness of planning and decision-making.
 Therefore, it is essential to develop real-time traffic planning and decision-making algorithms based on large-scale language models. This entails exploring how to combine the real-time traffic perception capabilities of generative AI technology with the decision-making capabilities of RL. Such an approach holds the potential to yield more efficient and adaptable solutions to tackle the intricacies and uncertainties inherent in traffic management.
As an illustrative example, consider driving decision-making in autonomous vehicles. Vehicles need to make decisions such as overtaking, decelerating, and stopping based on perceived data and environmental conditions. Large-scale models, equipped with powerful knowledge reasoning and generation capabilities, are expected to achieve precise traffic decisions in complex traffic environments. Therefore, future research can focus on how to design algorithms based on large-scale language models to make accurate decisions while ensuring a balance in real-time performance and safety.

\section{Conclusions}

Generative AI technology is playing an increasingly influential role in ITS, particularly in perception, prediction, simulation, and decision-making. 
In this survey, we systematically investigate key generative AI solutions within ITS, and explore the challenges that remain unresolved in this field.
Our analysis identifies several valuable guidelines. First, designing specialized HPC allows for large-scale traffic simulations and optimizations with massive multi-modal data. Second, ensuring data reliability and system security requires addressing data sparsity and safeguarding against adversarial attacks. Third, enhancing real-time performance while improving model interpretability is crucial to enable prompt responses to dynamic traffic conditions and foster trust through transparent decision-making.
Based on these guidelines, we summarize several key trends in generative AI for ITS. One significant trend is the integration of multi-modal data to enable comprehensive and large-scale traffic modeling. Another is the focus on designing generalizable models capable of addressing multiple transportation tasks simultaneously, which enhances the versatility and robustness of AI solutions. Additionally, there is growing interest in developing real-time traffic planning and decision-making algorithms, particularly by leveraging the capabilities of large-scale language models. These trends illustrate the transformative potential of generative AI in reshaping ITS, paving the way for safer, more efficient, and resilient transportation systems.

\bibliographystyle{ACM-Reference-Format}
\def\bibfont{\fontsize{5.5}{6.8}\selectfont}
\bibliography{sample-base.bib}

\end{document}